\documentclass[electronics,article,accept,pdftex,moreauthors]{Definitions/mdpi} 

\renewcommand\hl[1]{#1} 

\firstpage{1}
\pubvolume{13}
\issuenum{4}
\articlenumber{720}
\pubyear{2024}
\copyrightyear{2024}
\externaleditor{Academic Editor: Moez Bouchouicha \linebreak and Eric Moreau}
\datereceived{11 December 2023 } 
\daterevised{31 January 2024}
\dateaccepted{2 February 2024} 
\datepublished{9 February 2024} 

\hreflink{https://doi.org/10.3390/ electronics13040720}
\doinum{10.3390/electronics13040720}
\pdfoutput=1

\Title{Deep Spectral Meshes: Multi-Frequency Facial Mesh Processing with Graph Neural Networks}

\TitleCitation{Deep Spectral Meshes: Multi-Frequency Facial Mesh Processing with Graph Neural Networks}

\Author{Robert Kosk
 $^{1, 2,}$*\orcidA, Richard Southern $^{1}$\orcidB{}, Lihua You $^{1}$, Shaojun Bian $^{2,3}$\orcidD{}, Willem Kokke $^{2}$ and Greg Maguire $^{4}$\orcidE{}}

\AuthorNames{Robert Kosk, Richard Southern, Lihua You, Shaojun Bian, Willem Kokke and Greg Maguire}

\AuthorCitation{Kosk, R.; Southern, R.; \linebreak You, L.; Bian, S.; Kokke, W.; Maguire, G.}

\address{
$^{1}$ \quad \hl{Centre for Digital Entertainment, National Centre for Computer Animation}
, Bournemouth University, Poole, \hl{BH12 5BB, UK}
; \hl{rsouthern@bournemouth.ac.uk}
 (R.S.); \hl{lyou@bournemouth.ac.uk} (L.Y.)\\
$^{2}$ \quad Humain Ltd., \hl{Belfast, BT1 2LA}
, UK; \hl{shaojun@humain-studios.com (S.B.); willem@humain-studios.com (W.K.)}\\
$^{3}$ \quad School of Creative and Digital Industries, Buckinghamshire New University, \hl{High Wycombe, HP11 2JZ, UK}; \\
$^{4}$ \quad Belfast School of Art, Ulster University, \hl{Belfast, BT15 1ED, UK}; \hl{g.maguire@ulster.ac.uk}
}

\corres{Correspondence: rkosk@bournemouth.ac.uk}

\abstract{With the rising popularity of virtual worlds, the importance of data-driven parametric models of 3D meshes has grown rapidly. Numerous applications, such as computer vision, procedural generation, and mesh editing, vastly rely on these models. However, current approaches do not allow for independent editing of deformations at different frequency levels. They also do not benefit from representing deformations at different frequencies with dedicated representations, which would better expose their properties and improve the generated meshes' geometric and perceptual quality. In this work, spectral meshes are introduced as a method to decompose mesh deformations into low-frequency and high-frequency deformations. These features of low- and high-frequency deformations are used for representation learning with graph convolutional networks. A parametric model for 3D facial mesh synthesis is built upon the proposed framework, exposing user parameters that control disentangled high- and low-frequency deformations. Independent control of deformations at different frequencies and generation of plausible synthetic examples are mutually exclusive objectives. A Conditioning Factor is introduced to leverage these objectives. Our model takes further advantage of spectral partitioning by representing different frequency levels with disparate, more suitable representations. Low frequencies are represented with standardised Euclidean coordinates, and high frequencies with a normalised deformation representation (DR). This paper investigates applications of our proposed approach in mesh reconstruction, mesh interpolation, and multi-frequency editing. It is demonstrated that our method improves the overall quality of generated meshes on most datasets when considering both the $L_1$ norm and perceptual Dihedral Angle Mesh Error (DAME) metrics.}


\begin{document}

\section{Introduction}
\label{sec:introduction}
The importance of generating high-fidelity digital humans is more relevant now than ever before. Not only are we seeing significant investments in new visual effects-intensive creative content from streaming services, but platforms like \textit{Meta} are also reported to have invested at least USD 36 billion in developing content and tools to populate the next-generation social media platforms based on Virtual Reality. The availability of high-quality facial scans and recent advances in deep learning methods applied to mesh processing have led to the development of data-driven parametric models. These approaches are at the forefront of content creation technologies, allowing artists and users to rapidly generate high-fidelity character assets for these applications. However, parameters exposed by deep learning approaches may not always be meaningful. Furthermore, the perceptual and geometric quality of assets generated by neural models often does not meet standards appropriate for industrial applications.

The above issues can be addressed with suitable representations, which can describe deformations at different frequencies and independently edit deformations at these different frequency levels. This topic has not been investigated in existing work, as discussed below.

Polygon meshes \cite{Russo2010}, including quad and triangle meshes, are the most popular surface representation and are widely applied in the generation of creative content. They represent 3D surfaces with geometric vertices as absolute coordinates, making it challenging to edit overall shape deformations while preserving local surface details. In contrast, differential coordinates \cite{Feng&Shi2009}, often called the Laplacian operator or Laplacian coordinates, explicitly describe local surface and enable global shape editing while preserving local deformations~\cite{sorkine2004laplacian}. Since differential coordinates are only translation-invariant and not scale- and rotation-invariant, some improvements have been made to extend differential coordinates with these additional properties, as reviewed in Section \ref{subsec:differential coordinates-based 3D Shape Representation}. Other mesh operators \cite{Zhang2010SpectralProcessing} have also been introduced to extend the Laplacian operator. The Laplace–Beltrami operator, a generalisation of the Laplace operator, is used in this paper to represent \mbox{mesh deformations.}

Although mesh operators enable global shape editing while preserving local deformations, they still cannot represent deformations at different frequencies or independently edit deformations at different frequency levels. Spectral mesh processing \cite{Sorkine2005} derives eigenvalues, eigenvectors, or eigenspace projections from the mesh operators and uses them to carry out desired tasks. It provides a powerful means to achieve different approximations of a 3D mesh with different frequencies. In this paper, spectral mesh processing is used to decompose mesh deformations into low- and high-frequency deformations.   

Three-dimensional meshes are non-Euclidean data, unlike Euclidean data such as voxels, which have an underlying grid structure and can be treated by extending already-existing 2D deep learning paradigms. The lack of grid structure poses a challenge when attempting to apply classical deep learning techniques to non-Euclidean data. To address this problem, geometric deep learning \cite{Bronstein2017} has been developed explicitly for non-Euclidean data. Consequently, geometric deep learning is used in the proposed method to learn the latent parameters of low- and high-frequency deformations.

Parametric models, such as 3D morphable models \cite{Egger2020}, are commonly used to synthesise new meshes by altering the coefficients in a parametric space. They are widely applied due to their ability to model intrinsic properties of 3D faces. As reviewed in Section~\ref{subsec:Parametric_Models_with_Graph_Networks}, parametric models have been used in graph neural networks to represent facial shapes. A parametric model is also used in this paper for 3D facial mesh synthesis.

As the most popular 3D structure for representing 3D models, triangle meshes are graphs. Graph neural networks are suitable for the geometric learning of triangle meshes~\cite{Xiao2020}. Therefore, triangle meshes are considered in our proposed methods, and graph neural networks are used for deep learning.

By integrating the above-discussed Laplace–Beltrami operator for deformation representation, spectral mesh processing for decomposition of mesh deformations, geometric deep learning, and parametric models with graph neural networks, a new 3D facial mesh synthesis model is developed in this paper. The model exposes user parameters to control disentangled low- and high-frequency deformations, generate plausible facial shapes, and allow the user to control deformations independently at low- and \mbox{high-frequency levels.}

Generating plausible facial shapes and allowing controllable deformation can conflict, as plausible faces exist within a joint distribution of low- and high-frequency information. For example, wrinkled skin at higher frequencies correlates with volume loss of fat pads at lower frequencies. This problem is exacerbated when using smaller or biased datasets, which can lead to undesirable correlations. A Conditioning Factor is introduced to overcome the conflict between plausibility and user control. It is a scalar that modulates the influence of mutual conditioning of low- and high-frequency deformations. 

\textls[-9]{Lower-frequency deformations encode most of the mesh volume, while higher-frequency} deformations describe fine surface details. This observation is used to improve the quality of generated meshes. Low frequencies are represented with standardised Euclidean coordinates, which capture first-order mesh properties. The highest frequency deformation is represented with normalised deformation representation (DR) to improve perceptual quality. In this way, our proposed method improves the overall quality of generated meshes from geometric and perceptual perspectives, as shown in Section~\ref{sec:results}.

Our main contributions are:
\begin{itemize}[leftmargin=*, labelsep=4.9mm] 
\item The introduction of spectral decomposition of meshes in 3D shape representation learning.
\item A novel parametric deep face model which enables independent control of high- and low-frequency deformations.
\item Enhanced geometric and perceptual quality of generated meshes, achieved through the use of different representations of deformations at high and low frequencies.
\end{itemize}

The remaining parts of this paper are organised as follows. Related work is reviewed in Section \ref{sec:related_work}. Section \ref{sec:overview} offers an overview of the proposed approach. Deep spectral meshes are described in Section \ref{sec:DSM}. The impacts of the Conditioning Factor are discussed in Section \ref{sec:conditioning_influence}. The implementation of the proposed approach is detailed in Section \ref{sec:implementation}. Section \ref{sec:results} investigates applications and comparisons. Finally, Section \ref{sec:conclusions} provides a summary and identifies future work directions.

\section{Related Work}
\label{sec:related_work}
The work described in this paper is related to differential coordinate-based 3D shape representations, spectral mesh processing, geometric deep learning in the spectral domain, and parametric models with graph networks. This section offers a brief review of the existing work in these fields.

\subsection{Differential Coordinate-Based 3D Shape Representation} \label{subsec:differential coordinates-based 3D Shape Representation}
A translation-invariant differential representation for surface reconstruction is reviewed in \cite{Sorkine2005}, which uses Laplacian coordinates, also called differential coordinates, to represent surfaces. Laplacian coordinates allow for mesh editing and shape approximation. The paper also reviews the work of extending the translation-invariant differential representation to rotation-invariant differential representations. 

Gao et al. propose a new differential representation called rotation-invariant mesh difference (RIMD) \cite{Gao2016}. Unlike the translation-invariant differential representation, it encodes deformations and is invariant to rigid rotations and translations. RIMD is defined on both vertices and edges. Therefore, it is incompatible with our method, which requires a vertex-based representation. Gao et al. also propose an as-consistent-as-possible (ACAP) \cite{Gao2019} deformation representation as an improvement of computational issues of RIMD and its ability to handle large rotations. This deformation representation has also been used \mbox{in \cite{Tan2022}} to develop a mesh-based variational autoencoder architecture for dealing with meshes with irregular connectivity and non-linear deformations. 

The deformation representation proposed in \cite{Gao2019} requires the orientations of rotation axes and rotation angles of adjacent vertices to be as consistent as possible. However, in the context of facial deformations, local rotations never exceed $2\pi$ rad. Wu et al. address this and use less complex deformation representation (DR) based on the deformation gradient. DR has been successful in reconstructing caricatures of faces through the combination and extrapolation of deformation features \cite{Wu2018}.

Quantised DR is chosen to represent high-frequency deformations in our proposed approach. Similarly to other differential representations, quantised DR preserves local surface properties. Moreover, it is compatible with our method, which requires the features to be defined on vertices.

\subsection{Spectral Mesh Processing} \label{subsec:spectral_mesh_processing}
Various methods of spectral mesh processing have been developed to address the problems of shape analysis, mesh simplification, surface segmentation, and shape correspondence. Sorkine \cite{Sorkine2005} and Zhang et al. \cite{Zhang2010SpectralProcessing} comprehensively review existing work until 2005 and 2010, respectively. Recent work in the field is reviewed in the following paragraphs.

Melzi et al. obtain smooth, local, controllable, orthogonal, and efficient basis called Localized Manifold Harmonics (LMH) through the spectral decomposition of new types of intrinsic operators. The authors integrate local details provided by the basis and the global information from the Laplacian eigenfunctions to deal with local spectral shape \mbox{analysis \cite{Melzi2018}.} Xu et al. fit the original 3D model with a finite subdivision surface and restrict the eigenproblem with a subdivision linear subspace to obtain intrinsic shape information of 3D models through fast calculation of large-scale Laplace--Beltrami eigenproblem on the models \cite{Xu2021}. Lescoat et al. propose a new mesh simplification algorithm, which uses a spectrum-preserving mesh decimation scheme to simplify input meshes and makes the Laplacian of simplified meshes spectrally close to the Laplacian of input meshes \cite{Lescoat2020}.

Wang et al. use a Laplacian operator and select and combine some of its sub-eigenvectors to compute a single segmentation field to conduct mesh segmentation \cite{Wang2014}. Tong et al. build a Laplacian matrix, propose a spectral mesh segmentation method, which converts mesh segmentation into an $ \ell_0$ gradient minimisation problem, and devise a fast algorithm to solve the minimisation problem \cite{Tong2020}. Bao et al. devise a feature-aware simplification algorithm to create a coarse mesh. Then, they use the spectral segmentation method proposed in \cite{Tong2020} to perform partition on the coarse mesh to obtain a coarse segmentation. Next, they reverse the simplification process to map the coarse mesh to the input mesh and smooth jaggy boundaries to develop a spectral segmentation method for large meshes \cite{Bao2023}.

Jain and Zhang first transform two 3D meshes into spectral domain and find and match the embeddings of the two 3D meshes in the spectral domain to obtain vertex-to-vertex correspondence between the two 3D meshes \cite{Jain2006}. Dubrovina and Kimmel use the eigenfunctions of the Laplace--Beltrami operator to calculate surface descriptors and match surface descriptors to develop a correspondence detection method \cite{Dubrovina2010}. Melzi et al. introduce iterative spectral upsampling to obtain high-quality correspondences with a small number of coefficients in the spectral domain \cite{Melzi2019}.

Spectral methods share a common framework. They define a matrix representing a discretisation of a continuous operator over a mesh. In our case, it is the Laplace--Beltrami operator defined in Equation (\ref{eqn:laplacian}). Then, the matrix is decomposed to obtain eigenvectors and eigenvalues. These are used to solve problems in a specific domain. This paper introduces the concept of spectral mesh processing to 3D shape representation learning. Through the spectral decomposition of meshes, our method benefits from the ability to use different representations for low- and high-frequency data.

\subsection{Geometric Deep Learning in Spectral Domain} \label{subsec:geometric_deep_learning}
Deep neural networks have proven to be powerful tools in extracting latent representations of data in the Euclidean domain, especially 2D images. In recent years, these deep learning approaches extended to irregular graphs \cite{Bronstein2017,Wu2020,Xiao2020} and allowed for learning non-linear, lower-dimensional embeddings of meshes, including human faces. There are many publications on geometric deep learning. The current research on geometric deep learning in the spectral domain is briefly reviewed in the following paragraph.

Dong et al. propose a convolutional neural network framework called Laplacian2Mesh by mapping 3D meshes to a multi-dimensional Laplacian--Beltrami space to deal with irregular triangle meshes for shape classification and segmentation \cite{Dong2023}. Lemeunier et al. integrate spectral mesh processing and deep learning models consisting of a convolutional autoencoder and a transformer to develop a spectral transformer called SpecTrHuMS to generate human mesh sequences \cite{Lemeunier2023}. Qiao et al. present a deep learning approach that uses Laplacian spectral clustering to build a fine-to-coarse mesh hierarchy and integrate Laplacian spectral analysis and mesh feature aggregation blocks to encode mesh connectivity for shape segmentation and classification \cite{Qiao2022}.  
Based on the idea of approximating low and mid frequencies on coarse grids, Nasikun and Hildebrandt investigate a new solver called the Hierarchical Subspace Iteration Method that can solve sparse Laplace–Beltrami eigenproblems on meshes faster than existing methods based on Lanczos iterations, preconditioned conjugate gradients, and subspace iterations \cite{Nasikun2022}.  

\subsection{Parametric Models with Graph Networks} \label{subsec:Parametric_Models_with_Graph_Networks}

Work \cite{Ranjan2018} is the first geometric learning approach to representing facial shapes. It utilises a variational autoencoder architecture with spectral graph convolutional operations \cite{Defferrard2016}. Models \cite{Bouritsas2019NeuralGeneration,Chen2021LearningModels,Gao2021LearningRepresentation,Zhou2020FullyKernels,Verma2018} improve this architecture with custom convolutional and aggregation operators. Cheng et al. introduce MeshGAN, the adversarial approach to learning the facial manifold of 3D meshes \cite{Cheng2019}. Zhou et al. \cite{Zhou2019DenseDecoders} jointly learn deformation offset and colour residing on vertices of a face mesh \cite{Zhou2019DenseDecoders}. Yuan et al. convolve preprocessed ACAP deformation features \cite{Yuan2019}. Jiang et al. learn disentangled manifolds of facial identities and expressions in the normalised deformation representation (DR) \cite{Jiang2019}. Based on the deformation representation proposed in \cite{Gao2019}, a new variational autoencoder architecture is proposed in \cite{Zheng2021} to learn a latent representation of 3D human hands for high-accuracy monocular RGB-D/RGB 3D hand reconstruction.

The decomposition proposed in this paper is deformation decomposition, which decomposes a deformation into low- and high-frequency parts. It differs from the attribute decomposition proposed in \cite{Jiang2019}, where a 3D face shape is decomposed into identity and expression parts. However, both this paper and \cite{Jiang2019} adopt the same deformation representation proposed in \cite{Gao2019}. The deep spectral mesh graph neural network in our proposed method also differs from the variational autoencoder architecture in \cite{Zheng2021}. While \cite{Zheng2021} encodes and decodes ACAP deformation features using fully connected layers, our proposed approach separately encodes the decomposed deformation using graph convolutional layers and subsequently decodes the decomposed deformation from concatenated latent codes.

As discussed above, none of the parametric models allow for independent editing of deformations at different frequencies. Additionally, these models represent a whole spectrum of deformations with only one representation. Therefore, unlike our proposed approach, they cannot benefit from defining deformations at several frequency bands with different, dedicated representations.

\section{Method Overview}
\label{sec:overview}

\begin{figure}[H]
  \includegraphics[width=\textwidth]{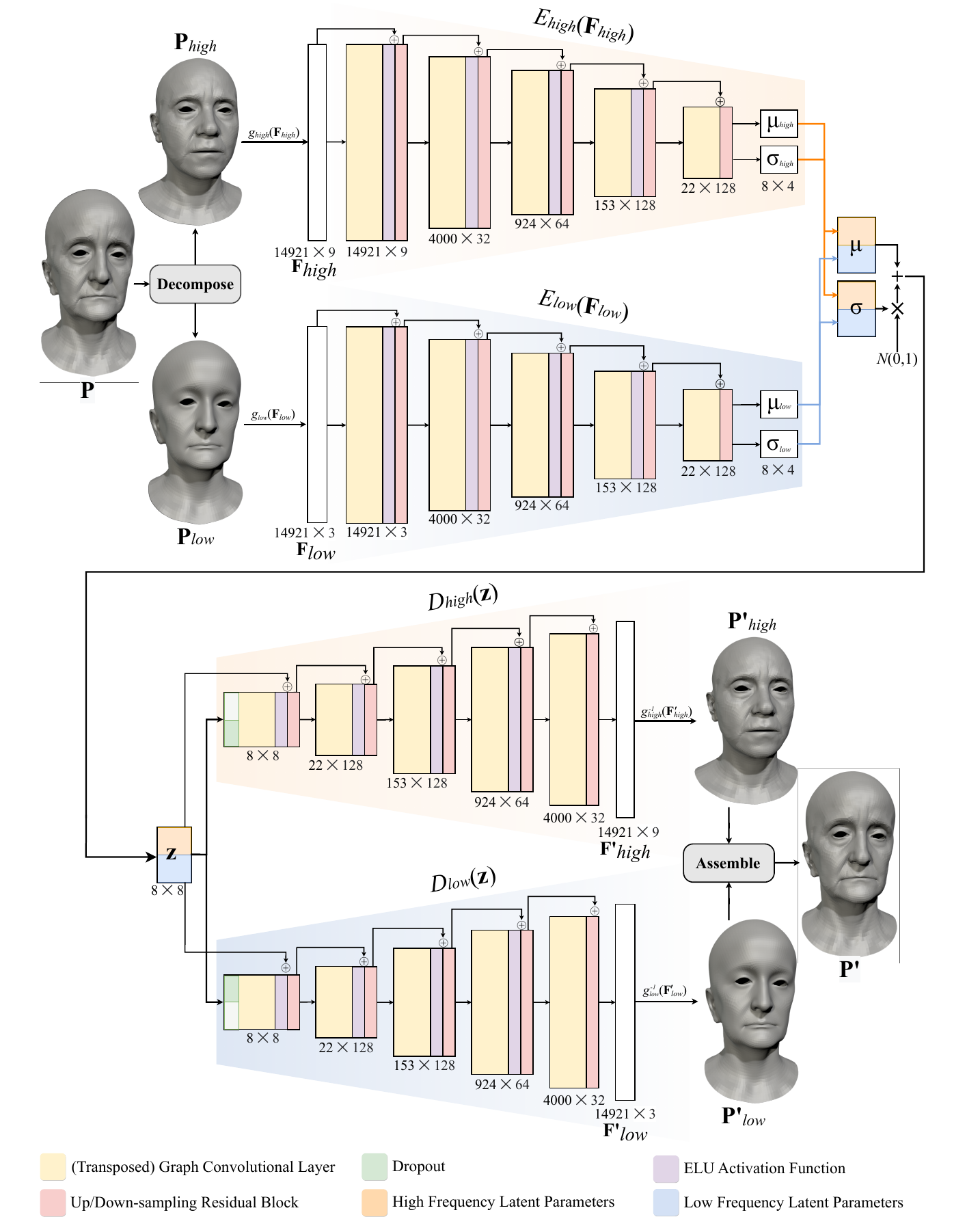}

  \caption{\label{fig:network_diagram}
          Overview of our Deep Spectral Meshes graph neural network. }
\end{figure}

Our proposed method is outlined in Figure \ref{fig:network_diagram}. It consists of three steps: spectral mesh processing,  neural networks, and final assembly. 

In the first step, preprocessed meshes $\mathbf{P}$ 
 are partitioned into two frequency bands through spectral decomposition. The resulting low- and high-frequency deformations $\mathbf{P}_{low}$ and $\mathbf{P}_{high}$ are transformed to standardised Euclidean coordinate features $\mathbf{F}_{low}$ and normalised deformation representation (DR) features $\mathbf{F}_{high}$. The spectral decomposition and representations of features are discussed in Section \ref{subsec:spectral_partitioning} below.

In the second step, a neural network consisting of graph encoders and graph decoders is used to reconstruct input features. Graph encoders $E_{high}$ and $E_{low}$ encode the features to latent means $\boldsymbol{\mu}_{high}$ and $\boldsymbol{\mu}_{low}$ and deviations $\boldsymbol{\sigma}_{high}$ and $\boldsymbol{\sigma}_{low}$. Means and deviations are concatenated, and latent codes $\mathbf{Z}$ are sampled from the distribution ([$\boldsymbol{\mu}_{high}~\mid~\boldsymbol{\mu}_{low}$],[$\boldsymbol{\sigma}_{high}~\mid~\boldsymbol{\sigma}_{low}$]). Graph decoders $D_{high}$ and $D_{low}$ reconstruct input features from $\mathbf{Z}$ and output reconstructed features $\mathbf{F}'_{high}$ and $\mathbf{F}'_{low}$ . The neural network is introduced in Section \ref{subsec:neural_network}. 

In the third step,  output features $\mathbf{F}'_{high}$ and $\mathbf{F}'_{low}$  are converted back from their representations to Euclidean coordinates $\mathbf{P}'_{high}$ and $\mathbf{P}'_{low}$, which are later combined to obtain the final vertex positions, $\mathbf{P}'$, as described in Section \ref{subsec:final_assembly}.

\subsection{Spectral Partitioning and Representation} \label{subsec:spectral_partitioning}
The generalised spectral decomposition \cite{Zhang2010SpectralProcessing} of a Laplacian matrix, $\mathbf{L}$, has a form of
\begin{equation}
    \mathbf{L} = \mathbf{U} \Lambda \mathbf{U}^T \mathbf{M},
    \label{eqn:eigenvalue_problem}
\end{equation}
where $\mathbf{M}$ is the mass matrix representing Voronoi areas around vertices. The use of mass matrix $\mathbf{M}$ counters the impact of irregular tessellation and, in consequence, improves the geometric quality of meshes generated by our method.

Spectral partitioning is the process of identifying $n$ subsets of eigenvectors $\mathbf{U}$ and projecting vertex positions $\mathbf{P}$ of a mesh onto these subsets of vectors to isolate different frequency bands $\{\mathbf{P}_0, ..., \mathbf{P}_n\}, \mathbf{P}=\sum_{i=0}^{n}\mathbf{P}_i$. The choice of size and number of frequency bands is arbitrary and depends on a specific application. Projection of $\mathbf{P}$ onto a subset of eigenvectors $\mathbf{U}_i$ is defined as
\begin{equation}
     \mathbf{P}_i = \mathbf{U}_i \mathbf{U}^T_i \mathbf{M} \mathbf{P}.
\end{equation}

As a consequence of spectral partitioning, frequency bands $\{\mathbf{P}_0, ..., \mathbf{P}_n\}$ can be transformed into different, dedicated representations, which are more suitable for a given spectral band. 
A representation transform can be denoted as function $g_i(\mathbf{P}_i)$, which transforms the $i$th frequency band into representation $\mathbf{F}_i$. An associated inverse transform can be denoted as $g^{-1}_i(\mathbf{F}_i)$. These functions must output a per-vertex representation because $\mathbf{F}_i$ are inputs to graph neural networks which process input signal defined on vertices.

\subsection{Neural Network} \label{subsec:neural_network}
Variational graph autoencoders \cite{Zhou2020FullyKernels} are used to encode features of each frequency band into their respective latent distributions $\mathcal{N}(\boldsymbol{\mu}_{i}, \boldsymbol{\sigma}_{i}$), where $\boldsymbol{\mu}_{i}$ is a mean vector and $\boldsymbol{\sigma}_{i}$ is a standard deviation vector. If desired, each encoder can output a different number of parameters. The optimal size of the parametric space depends on the training dataset and requirements of a specific application. Therefore, it can be determined as part of the hyperparameter optimisation process.

The neural network consists of $n$ graph encoders $E_i$ with $n$ corresponding decoders $D_i$. The choice of convolutional, transpose convolutional, pooling, de-pooling or other layers of encoders and decoders is arbitrary. While each encoder outputs a different latent distribution of its frequency band, the encoders take the same set of parameters as input. Specifically, mean vectors are concatenated to  $\boldsymbol{\mu} = [\boldsymbol{\mu}_0, ..., \boldsymbol{\mu}_n]$ and deviation vectors are concatenated to $\boldsymbol{\sigma} = [\boldsymbol{\sigma}_0, ..., \boldsymbol{\sigma}_n]$. Afterwards, latent parameters $\mathbf{Z}$ are stochastically sampled from $\mathcal{N}(\boldsymbol{\mu}, \boldsymbol{\sigma}$). The decoders are optimised to output reconstructed features $\mathbf{F}'_i$, with the objective of minimising the discrepancy between $\mathbf{F}_i$ and $\mathbf{F}'_i$.

\subsection{Final Assembly} \label{subsec:final_assembly}
Before combining reconstructed vectors $\mathbf{F}'_i$, it is necessary to convert them back to Euclidean coordinates so that the reconstructed frequency band $\mathbf{P}'_i = g^{-1}_i(\mathbf{F}'_i)$. It can be observed that each decoder $D_{i}$ is optimised to generate outputs which, after transformation with $g^{-1}_i(\cdot)$, are a linear combination of eigenvectors $\mathbf{U}_i$. This implies that $\mathbf{P}'_i$ can be projected onto these vectors without information loss, as the filtered-out frequencies are noise which lies outside the domain for this frequency band. Based on this observation, the final reconstructed vertex positions are obtained through the aggregation of frequency bands $\mathbf{P}'_{i}$ in the following way:
\begin{equation} \label{eqn:combined_outputs2}
    \mathbf{P}' = \sum_{i=0}^{n}\mathbf{U}_i \mathbf{U}^T_i \mathbf{M} \mathbf{P}'_i \quad.
\end{equation}

\section{Deep Spectral Meshes}
\label{sec:DSM}

This section explains the details of our parametric deep face model, which uses the spectral decomposition of meshes. An overview of the model is provided in Figure~\ref{fig:network_diagram}.

\subsection{Vertex Feature Representation} \label{sec:vertex_feature_representation}

Given a dataset of triangle meshes with shared connectivity, i.e., with the same triangle-vertex index incidence table to associate each triangle with its three bounding vertices, the triangle meshes are translated to make their centre coincide with the origin. Then, their mean is computed, which is an averaged shape of all triangle meshes denoted as $\bar{\mathbf{P}}$. Next, each triangle mesh is translated and rotated to align with the mean. Following this, the triangle meshes are uniformly scaled to fit within a cube with sides measuring two units each.

\subsubsection{Spectral Decomposition} \label{sec:spectral_decomposition}

\begin{figure}[H]%

\includegraphics[width=0.5\textwidth]{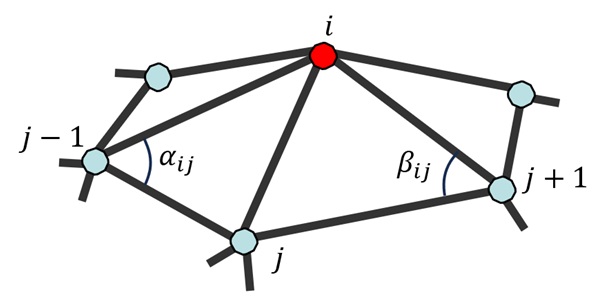}
\caption{One-ring neighbourhood vertices and the angles used to calculate cotangent weights.}\label{one_ring_n}
\end{figure}

 Laplacian matrix $\mathbf{L}$ in Equation (\ref{eqn:eigenvalue_problem}) consists of elements $\mathbf{L}_{ij}$, which are determined by the following cotangent Laplacian operator used in~\cite{Feng&Shi2009}:
\begin{equation} \label{eqn:laplacian}
  \mathbf{L}_{ij} =
    \begin{cases}
      j \in \mathcal{N}_i & \cot \alpha_{ij} + \cot \beta_{ij}\\
      j \not\in \mathcal{N}_i & 0\\
      i = j & - \sum_{k \neq j} \mathbf{L}_{ik}
    \end{cases},
\end{equation}
where $\mathcal{N}_i$ are vertices ..., $j-1$, $j$, $j+1$, ... in 
the one-ring neighbourhood of vertex \mbox{$i$ and $\alpha_{ij}$} and $\beta_{ij}$ are the angles opposite to edge $ij$, as shown in Figure~\ref{one_ring_n}.

For a large triangle mesh, Laplacian matrix $\mathbf{L}$ in Equation (\ref{eqn:eigenvalue_problem}) has many eigenvectors. Calculating all the eigenvectors of such Laplacian matrix is slow. To tackle this problem, only the first $k$ eigenvectors are calculated. Meshes are partitioned into two frequency bands: low and high frequencies. The former band is defined by the first $k$ eigenvectors $\mathbf{U}_{(k)}$, which correspond to the k smallest eigenvalues of Laplacian matrix $\mathbf{L}$. Section \ref{subsec:high_frequency_band} covers the high-frequency band.

We let $\hat{\mathbf{P}}$ be a deformation from a mean so that $\hat{\mathbf{P}} = \mathbf{P} - \bar{\mathbf{P}}$. As eigenvectors $\mathbf{U}_{(k)}$ correspond to the lowest eigenvalues, they form a discrete space of slowly changing values. Therefore, by projecting $\hat{\mathbf{P}}$ onto space $\mathbf{U}_{(k)}$, the resulting vertex positions \mbox{$\mathbf{P}_{\text{low}}$ are $\hat{\mathbf{P}}$}, which is a mean deformed by only low-frequency deformations. $\mathbf{P}_{\text{high}}$ represents the remaining high-frequency deformations of a mean $\bar{\mathbf{P}}$. Meshes $\mathbf{P}_{\text{low}}$ and $\mathbf{P}_{\text{high}}$ are computed as follows:
\begin{equation}
\begin{aligned}
    \mathbf{X} = \mathbf{U}_{(k)} \mathbf{U}^T_{(k)} \mathbf{M}, \\
    \mathbf{P}_{\text{low}} = \mathbf{X}\hat{\mathbf{P}} + \bar{\mathbf{P}}, \\
    \mathbf{P}_{\text{high}} = (\mathbf{I}-\mathbf{X})\hat{\mathbf{P}} + \bar{\mathbf{P}} .
\end{aligned}
\label{eqn:decomposition}
\end{equation}

\subsubsection{High-Frequency Band} \label{subsec:high_frequency_band}

High-frequency band $\mathbf{P}_{high}$ is encoded in a normalised deformation representation (DR) \cite{Gao2019,Wu2018}. This representation encodes per-vertex deformation gradient $\mathbf{T}_i$ between the position of vertex $\mathbf{p}_i$ on mean $\bar{\mathbf{P}}$ and the position of deformed vertex \mbox{$\mathbf{p}'_i$ on $\mathbf{P}_{high}$.} Following \cite{Baran2009,Sorkine2007}, deformation gradient $\mathbf{T}_i$ is calculated by solving the following weighted least-squares system, which minimises energy $E(\mathbf{T}_i)$ such that
\begin{equation} \label{eqn:def_gradient}
E(\mathbf{T}_i) = \sum\limits_{j \in \mathcal{N}_i} c_{ij} \bigg\lVert(\mathbf{p}'_i - \mathbf{p}'_j) - \mathbf{T}_i (\mathbf{p}_i - \mathbf{p}_j) \bigg\rVert^2 \ \ ,
\end{equation}
where $c_{ij}$ are cotangent weights calculated on a mean of the training meshes.

As linear interpolation of matrices $\mathbf{T}_i$ is meaningless, they are decomposed into a rotational part and a scale/shear part using polar decomposition so that $\mathbf{T}_i = \mathbf{R}_i \mathbf{S}_i$. The rotation matrix $\mathbf{R}_i$, mapped to $\log \mathbf{R}_i$, can be linearly interpolated and then converted back to $\mathbf{R}_i = \exp(\log \mathbf{R}_i)$. Finally, identity matrix $\mathbf{I}$ is subtracted from $\mathbf{S}_i$.

Although both per-vertex matrices $\mathbf{R}_i$ and $\mathbf{S}_i$ contain nine elements each, only six non-trivial scale elements and three non-trivial rotation elements are combined to construct deformation features $\mathbf{f}_i$, where $|\mathbf{f}_i|=9$. The results of our experiments, shown in Table~\ref{table:impact_of_normalisation_and_standardisation}, demonstrate that the normalisation of DR results in a lower reconstruction error of test data. Consequently, each channel is normalised to range $[-1,1]$.

\subsubsection{Low-Frequency Band} \label{subsec:low_frequency_band}
Meshes $\mathbf{P}_{\text{low}}$ remain in the Euclidean coordinate representation. Based on our experiments in Section \ref{sec:results}, inputs $\mathbf{F}_{low}$ are standardised by subtraction of the mean and division by standard deviation.

\subsection{Graph Network Architecture} \label{subsec:graph_network_architecture}
Our network consists of two fully convolutional variational graph autoencoders, as depicted in Figure~\ref{fig:network_diagram}. Encoders $E_{high}$ and $E_{low}$ take as input high-frequency features $\mathbf{F}_{high}$ and low-frequency features $\mathbf{F}_{low}$, respectively (see Section \ref{sec:vertex_feature_representation}). The encoders compress each input feature to compact distributions with means $\boldsymbol{\mu}_{high}$ and $\boldsymbol{\mu}_{low} \in \mathbb{R}^{8 \times 4}$, and deviation vectors $\boldsymbol{\sigma}_{high}$ and $\boldsymbol{\sigma}_{low} \in \mathbb{R}^{8 \times 4}$. Subsequently, output means and deviations are concatenated so that $\boldsymbol{\mu} = [\boldsymbol{\mu}_{high} \mid \boldsymbol{\mu}_{low}]$ and $\boldsymbol{\sigma} = [\boldsymbol{\sigma}_{high} \mid \boldsymbol{\sigma}_{low}]$.

Latent code $\mathbf{Z} \in \mathbb{R}^{8 \times 8}$ is obtained in a stochastic process. First, standard deviation $\boldsymbol{\sigma}$ is multiplied with scalar $\boldsymbol{\epsilon}$ drawn from a normal distribution. Then, the result is added to mean vector $\boldsymbol{\mu}$ so that $\boldsymbol{\sigma} \times \boldsymbol{\epsilon} + \boldsymbol{\mu} = \mathbf{Z} = [\mathbf{Z}_{high} \mid \mathbf{Z}_{low}]$.

Both decoders, $E_{high}$ and $E_{low}$, take $\mathbf{Z}$ as input. This way, $E_{high}$ decodes high-frequency deformation features $\mathbf{F}'_{high}$ from code $\mathbf{Z}_{high}$ conditioned on code $\mathbf{Z}_{low}$. Analogically, $E_{low}$ decodes low-frequency deformation features $\mathbf{F}'_{low}$ from code $\mathbf{Z}_{low}$ conditioned on code $\mathbf{Z}_{high}$. Controlling the influence of this conditioning is discussed in Section \ref{sec:conditioning_influence}.

\subsection{Network Structure}
Both variational graph autoencoders have the same structure. They use convolution/transposed convolution operations and upsampling/downsampling residual layers introduced in \cite{Zhou2020FullyKernels}. However, other types of graph convolutional and pooling operators can be used instead if they accept input features defined solely on vertices. Therefore, the encoders and decoders operating on signal defined on edges would not be suitable, as our method uses low- and high-frequency features defined on vertices.

The encoders put their inputs through five graph convolutional layers. The local convolution kernels are sampled from the global kernel weight basis to perform convolutional operations on irregular graphs. Therefore, the network learns the shared global kernel weight basis and per-vertex sampling functions \cite{Zhou2020FullyKernels}. We use stride~=~2, kernel radius~=~2, weight basis = 35, and channel dimensions = [$|\mathbf{f}|$, 32, 62, 128, 128, 4]. All layers, except the last one, are followed by the $\textit{ELU}$ activation function. The outputs from the $\textit{ELU}$ are added to outputs from a downsampling residual layer. The decoders mirror the encoders with five blocks of graph transposed convolutional layers followed by the $\textit{ELU}$s and upsampling residual layers.

\subsection{Training Process} \label{subsec:training_process}
The two variational graph autoencoders are trained simultaneously. At each iteration, the weights and biases of each encoder--decoder pair are updated through backpropagation using two separate Adam optimisers. Learnable parameters of $E_{high}$ and $D_{high}$ are optimised in terms of loss $L_{high}$, while parameters of $E_{low}$ and $D_{low}$ are updated in terms of loss $L_{low}$. Losses are calculated as follows:
\begin{equation} \label{eqn:loss_function_high}
\begin{aligned}
L_{high} = ||\mathsf{denorm}(\mathbf{F}_{high}) - \mathsf{denorm}(\mathit{D}_{high}(\mathbf{Z}))||_1 \\+ \phi \mathsf{KL}(\mathcal{N}(0,1)||p(\mathbf{Z}_{high}|\mathbf{F}_{high})),
\end{aligned}
\end{equation}
\begin{equation} \label{eqn:loss_function_low}
\begin{aligned}
L_{low} = ||\mathsf{destd}(\mathbf{F}_{low}) - \mathit{D}_{low}(\mathbf{Z})||_1 \\ + \phi \mathsf{KL}(\mathcal{N}(0,1)||p(\mathbf{Z}_{low}|\mathbf{F}_{low})) .
\end{aligned}
\end{equation}
In other words, both losses are a sum of two terms. The first one is the $L_1$ norm of a difference between the ground truth feature and its reconstruction. The second one is weighted Kullback–Leibler (KL) divergence, which measures a difference between normal distribution $\mathcal{N}(0,1)$ and latent vector distribution, where $\phi$ is a scalar weight.

Since $D_{high}$ outputs the normalised DR features (see Section \ref{subsec:high_frequency_band}), the output, $\mathbf{F}'_{high}$, as well as the ground truth, $\mathbf{F}_{high}$, are denormalised before calculating the $L_1$ norm of a difference between them. Denormalisation is denoted with the $\mathsf{denorm}(\cdot)$ function. In the case of $D_{low}$, the decoder outputs a destandardised feature in Euclidean coordinates. Therefore, only ground truth feature $\mathbf{F}_{low}$ is destandardised before computing the $L_1$ norm term. Here, destandardisation is denoted with the $\mathsf{destd}(\cdot)$ function.

\subsection{Inference} \label{subsec:inference}
This section describes the postprocessing steps required to obtain reconstructed vertex positions $\mathbf{P}'$. To be a valid input to the network, every mesh with vertex positions $\mathbf{P}$ must be represented with features $\mathbf{F}_{high}$ and $\mathbf{F}_{low}$, following the process described in Section~\ref{sec:vertex_feature_representation}. At inference, the network generates normalised DR features $\mathbf{F}'_{high}$ and Euclidean coordinates $\mathbf{F}'_{low} = \mathbf{P}'_{low}$. To calculate $\mathbf{P}'_{high}$, $\mathbf{F}'_{high}$ is denormalised and converted back from the normalised DR representation to Euclidean coordinate representation, as in \cite{Wu2018}. Subsequently, $\mathbf{P}'_{high}$ and $\mathbf{P}'_{low}$ are combined in the following way:
\begin{equation} \label{eqn:combined_outputs}
\begin{aligned}
    \mathbf{P}' = (\mathbf{I}-\mathbf{X}) \mathbf{P}'_{\textit{high}} + \mathbf{X} \mathbf{P}'_{\textit{low}},
\end{aligned}
\end{equation}
where $\mathbf{X}$ is the matrix from Equation (\ref{eqn:decomposition}).

\section{Conditioning Influence} \label{sec:conditioning_influence}
Conditioning, described in Section \ref{subsec:graph_network_architecture}, allows for the network to generate plausible facial meshes from a joint distribution of high- and low-frequency information. However, it can be desirable to tame the influence of conditioning to increase artistic control and prevent overfitting.

In an extreme case, the impact of conditioning could be removed if all the conditioning parameters were set to a constant value, for example, zero. On the other hand, if only a pseudo-randomly drawn fraction of these parameters was set to zero at each training iteration, the impact of conditioning would lower. It is proposed to achieve this effect by applying a dropout \cite{Hinton2012ImprovingDetectors} to the conditioning parameters.

Specifically, at each iteration, conditioning parameters $\mathbf{Z}_{low}$ of decoder $D_{high}$ are randomly zeroed with probability $(1 - \gamma)$ using the samples from the Bernoulli distribution. The same applies to conditioning parameters $\mathbf{Z}_{high}$ of decoder $D_{low}$. Therefore, scalar $\gamma$ can be called a Conditioning Factor, where $\gamma = 0$ prevents conditioning and $\gamma = 1$ means full conditioning.

\section{Implementation Details}\label{sec:implementation}

Our parametric models are trained with the following datasets of facial meshes: Facsimile~\texttrademark$ $~\cite{HumainLimited2022HumainDevelopment} (202 meshes, 14,921 vertices each), FaceScape \cite{Yang2020} (807 meshes, \mbox{26,317 vertices} each) and FaceWarehouse \cite{Cao2014} (150 meshes, 11,510 vertices each). As all three datasets contain subjects performing a set of facial expressions, only the shapes with a neutral expression are used. All meshes within each dataset have triangular faces with consistent connectivity. In our experiments, each dataset is split into training, validation and test subsets in proportions 85:5:10.

ARPACK \cite{LehoucqRichardBandSorensenDannyCandYang1998ARPACKMethods} is used to solve the generalised eigenvalue problem in Equation (\ref{eqn:eigenvalue_problem}). Only the first $k$ eigenvectors are computed. The networks are trained with $k=500$, the Conditioning Factor $\gamma = 1.0$ (full conditioning) and $\gamma = 0.4$ (partial conditioning).

The networks are implemented in Pytorch 1.8 \cite{PaszkeAdamandGrossSamandMassaFranciscoandLererAdamandBradburyJamesandChananGregoryandKilleenTrevorandLinZemingandGimelsheinNataliaandAntiga2019Pytorch:Library}. In all experiments, the networks are trained for 500 epochs, with a learning rate of $10^{-4}$ and a learning rate decay of 1\% after each epoch. KL divergence weight $\phi$ from Equations (\ref{eqn:loss_function_high}) and (\ref{eqn:loss_function_low}) is $10^{-6}$. Adam optimiser's hyper-parameters are set to $\beta_1=0.9$ and $\beta_2=0.999$. The networks trained with the Facsimile\texttrademark$ $ and FaceWarehouse datasets are optimised with a batch size of 16, while those trained with FaceScape have a batch size of 8 due to GPU memory constraints.

Matrix $\mathbf{X}$ from Equations (\ref{eqn:decomposition}) and (\ref{eqn:combined_outputs}) needs to be pre-computed only once. Otherwise, its frequent computation can significantly lower the performance of data preprocessing. Table \ref{table:performance} compares the CPU time and memory required to compute $\mathbf{U}$ and $\mathbf{X}$ on different datasets. Additionally, the table shows the CPU time of the final assembly Equation (\ref{eqn:combined_outputs}). This operation is performed each time a new mesh is inferred and takes 0.20 to 0.98 s per mesh, depending on the dataset used. For faster eigendecomposition of Laplacian $\mathbf{L}$ from Equation (\ref{eqn:laplacian}), approximation techniques can be used. Nevertheless, they are not used in our implementation.

\begin{table}[H]
\centering
\caption{\label{table:performance} Per-mesh CPU time and CPU memory required to compute terms from \mbox{Equations (\ref{eqn:decomposition}) and (\ref{eqn:combined_outputs}).} Three datasets of different vertex count are compared: FaceWarehouse \cite{Cao2014} (150 meshes, 11,510 verts), Facsimile~\texttrademark$ $~\cite{HumainLimited2022HumainDevelopment} (202 meshes, 14,921 verts) and FaceScape \cite{Yang2020} (26,317 verts).}
\resizebox{\textwidth}{!}{%
\begin{tabular}{@{}llll@{}}
\toprule
 &
  \begin{tabular}[c]{@{}l@{}}\textbf{FaceWarehouse} \\ \textbf{(11,510 verts)}\end{tabular} &
  \begin{tabular}[c]{@{}l@{}}\textbf{Facsimile\texttrademark$~$}\\ \textbf{(14,921 verts)}\end{tabular} &
  \begin{tabular}[c]{@{}l@{}}\textbf{FaceScape}\\ \textbf{(26,317 verts)}\end{tabular} \\ \midrule
Computation of $\mathbf{U}$ CPU time {[}s{]}          & 26.92 & 33.72 & 58.22 \\
Computation of $\mathbf{X}$ CPU time {[}s{]}          & 2.11  & 7.70  & 9.88  \\
Computation of Equation (\ref{eqn:combined_outputs}) CPU time {[}s{]} & 0.20  & 0.36  & 0.98  \\ \midrule
Computation of $\mathbf{U}$ CPU memory {[}MB{]}        & 45.0  & 58.3  & 102.8 \\
Computation of $\mathbf{X}$ CPU memory {[}GB{]}       & 1.04  & 1.74  & 5.41  \\ \bottomrule
\end{tabular}%
}
\end{table}

All the experiments are trained and inferred on a single NVIDIA GeForce GTX 1080 Ti with 16 GB memory. Data processing is performed on an Intel(R) Xeon(R) CPU E5-1630 v4 running at $8 \times 3.70$ GHz using 32 GB RAM. Training times range between 2 h for the FaceWarehouse dataset and 19 h for the FaceScape dataset.

\section{Applications and Comparisons} \label{sec:results}
The approach proposed in this paper has a large number of applications. Besides those identified in Section~\ref{sec:conclusions}, this section investigates the applications of the proposed approach in mesh reconstruction, mesh interpolation and multi-frequency mesh editing. It also compares the proposed approach with previously published methods.

\subsection{Mesh Reconstruction}

Mesh reconstruction can be used in representation learning, mesh compression, mesh smoothing and fairing. In this subsection, our proposed method is applied to reconstruct 3D meshes for mesh compression. For illustration, all the reconstruction experiments on the Facsimile\texttrademark$ $ dataset use a compression rate of 0.14\%. These reconstruction experiments compare our method and common representations used in other methods: Euclidean coordinates, standardised Euclidean coordinates and the normalised deformation representation (DR).

Implementation of our method follows the description from Sections \ref{sec:DSM} and \ref{sec:implementation}, with the Conditioning Factor $\gamma=1.0$ and frequencies split at $k=500$. All other representations do not decompose the mesh into deformations with multiple frequencies. Therefore, they are evaluated on the fully convolutional variational graph autoencoder with a single encoder and decoder. The encoder is the same as $E_{high}$ or $E_{low}$, and the decoder is the same as $D_{high}$ or $D_{low}$, without the dropout layer. All the experiments encode to latent space $\mathbf{Z}$ \mbox{of 64 parameters.}

\subsubsection{Quantitative Evaluation} \label{subsec:quantitative_evaluation}
Two metrics, the point-wise $L_1$ norm and the Dihedral Angle Mesh Error \mbox{(DAME) \cite{Vasa2012DihedralMeshes},} are used to quantitatively assess the quality of meshes generated by our method. The commonly used $L_1$ norm $ ||\mathbf{P} - \mathbf{P}'||_1$ between reconstructed Euclidean coordinates $\mathbf{P}'$ and the ground truth $\mathbf{P}$ is chosen due to its sensitivity to overall shape changes. In other words, the $L_1$ norm is capable of capturing low-frequency error. Nonetheless, it poorly correlates with high-frequency discrepancies perceived by the human visual system \cite{Corsini2013PerceptualMeshes}. Therefore, the results are also evaluated on a perceptual metric, as described below.

Given that the final meshes generated by our method are viewed by the human observer, the perceptual quality of the results is essential. Quantitative evaluation of perceptual mesh quality is often overlooked in previous work on parametric models with graph networks. In this work, DAME \cite{Vasa2012DihedralMeshes} is used to capture high-frequency discrepancies between the reconstructed and the ground truth meshes, and measure the perceptual quality of the generated results. DAME is selected due to its highest correlation with human judgement, measured on the compression task, compared to other common perceptual metrics. Moreover, DAME is dedicated to datasets with shared \mbox{connectivity \cite{Corsini2013PerceptualMeshes}.}

DAME consists of three elements: the difference between oriented dihedral angles, the masking effect and the visibility weighting. The last one is application-specific, as it changes with the viewing angles and the rendering resolution. Therefore, following the recommendation in \cite{Vasa2012DihedralMeshes}, the visibility term is replaced with triangle areas. Border edges are ignored in the calculation of DAME, as oriented dihedral angles cannot be calculated on these edges.

The reconstruction results generated with our approach are compared with those output by other methods: Mesh Autoencoder \cite{Zhou2020FullyKernels}, SpiralNet++ \cite{Gong2019SpiralNet++:Operator}, Neural 3DMM \cite{Bouritsas2019NeuralGeneration} and FeaStNet \cite{Verma2018}. Table~\ref{table:quantitative_comparison_to_other_methods} presents the quantitative outcomes of this comparison. Our proposed method consistently outperforms all counterparts in terms of the perceptual DAME metric across both the Facsimile\texttrademark$~$ and FaceWarehouse training and test datasets. Regarding point-wise accuracy measured with the $L_1$ norm, our method outperforms other compared methods on the FaceWarehouse training dataset. However, on the Facsimile\texttrademark$~$ training and test sets and the FaceWarehouse test dataset, our method performs less favourably than SpiralNet++ \cite{Gong2019SpiralNet++:Operator} and Neural 3DMM \cite{Bouritsas2019NeuralGeneration}. Nonetheless, a comprehensive assessment encompasses both perceptual and point-wise accuracy perspectives, and the perceptual results generated by \cite{Bouritsas2019NeuralGeneration,Gong2019SpiralNet++:Operator} have significantly higher perceptual DAME error. These quantitative results are further supported by a qualitative user study in Section~\ref{subsec:qualitative_evaluation}.

\begin{table}[H]

 \caption{\label{table:quantitative_comparison_to_other_methods} Numerical comparison of the reconstruction results of the Facsimile\texttrademark$ $ and FaceWarehouse datasets using our method ($k=500$, $\gamma=1$, $\mathbf{Z}=64$) and four other methods: Mesh Autoencoder \cite{Zhou2020FullyKernels}, SpiralNet++ \cite{Gong2019SpiralNet++:Operator}, Neural 3DMM \cite{Bouritsas2019NeuralGeneration} and FeaStNet \cite{Verma2018}.}
 \newcolumntype{C}{>{\centering\arraybackslash}X}
{%
\begin{tabularx}{\textwidth}{lCClCC} 
\toprule
\textbf{}     & \multicolumn{2}{c}{\textbf{Training}}       & \textbf{}            & \multicolumn{2}{c}{\textbf{Test}}           \\ \midrule
\textbf{} &
  \begin{tabular}[c]{@{}c@{}}\boldmath{${L_1}$} \textbf{Norm}\\ $\mathbf{\times10^{-3} \downarrow}$ \end{tabular} &
  \begin{tabular}[c]{@{}c@{}}\textbf{DAME}\\ $\mathbf{\times10^{-2} \downarrow}$\end{tabular} &
  &
  \begin{tabular}[c]{@{}c@{}}\boldmath{${L_1}$} \textbf{Norm}\\ $\mathbf{\times10^{-3} \downarrow}$\end{tabular} &
  \multicolumn{1}{l}{\begin{tabular}[c]{@{}l@{}}\textbf{DAME}\\ $\mathbf{\times10^{-2} \downarrow}$\end{tabular}} \\
  \midrule
\multicolumn{6}{l}{\textbf{Facsimile\texttrademark$ $}
}                                                                  \\ \midrule
Ours          & 1.61        & \textbf{2.76}        & \multicolumn{1}{c}{} & 6.42                 & \textbf{3.17}                 \\
Mesh Autoencoder.            & 2.60                 & 6.04                 & \multicolumn{1}{c}{} & 8.32                 & 5.81        \\
SpiralNet++ & \textbf{1.35}                 & 5.35                 & \multicolumn{1}{c}{} & 6.38        & 4.87                 \\
Neural 3DMM & 1.71                 & 3.84                 & \multicolumn{1}{c}{} & \textbf{5.95}        & 3.81                 \\
FeaStNet     & 2.02                 & 5.30                 & \multicolumn{1}{c}{} & 9.07                 & 5.35                 \\ \midrule
            
\multicolumn{6}{l}{\textbf{FaceWarehouse}}                                                             \\ \midrule
Ours          & \textbf{0.91}        & \textbf{1.10}        & \textbf{}            & 6.27                 & \textbf{1.29}                 \\
Mesh Autoencoder.            & 2.54                 & 5.27                 & \multicolumn{1}{c}{} & 5.33                 & 5.50        \\
SpiralNet++ & 1.21                 & 6.06                 & \multicolumn{1}{c}{} & 4.69        & 5.63                 \\
Neural 3DMM & 1.58                 & 4.84                 & \multicolumn{1}{c}{} & \textbf{4.02}        & 4.46                 \\
FeaStNet     & 1.92                 & 6.81                 & \multicolumn{1}{c}{} & 8.17                 & 6.30                 \\ \bottomrule
\end{tabularx}%
}
\end{table}

In Table \ref{table:quantitative_results}, the $L_1$ norm and DAME metrics are used to evaluate reconstructed meshes with our proposed approach and the 3D shape representations from other methods. It is demonstrated that, across the majority of datasets, our method outperforms other methods in reconstructing examples from the training set. Reconstruction of examples seen by the network during training has applications in 3D mesh compression.

\begin{table}[H]
 \caption{\label{table:quantitative_results} Quantitative comparison of the reconstruction results with our method ($k=500$, $\gamma=1$) and with common representations used in other methods: Euclidean coordinates \cite{Cheng2019,Hanocka2019MeshCNN:Edge,Zhou2020FullyKernels}, standardised Euclidean coordinates \cite{Bouritsas2019NeuralGeneration,Chen2021LearningModels,Gao2021LearningRepresentation,Gong2019SpiralNet++:Operator,Ranjan2018} and the normalised deformation representation (DR) \cite{Jiang2019,Wu2018}. To ensure a fair comparison between our method and other input representations, they are evaluated on the fully convolutional variational graph autoencoder with a single encoder and a single decoder. The encoder is the same as $E_{high}$ or $E_{low}$, and the decoder is the same as $D_{high}$ or $D_{low}$, without the dropout layer. All the comparisons encode to latent space $\mathbf{Z}$ of 64 parameters. Our method outperforms the reconstruction of examples from the training set on most datasets. On the test set, our method favourably compromises between the point-wise $L_1$ precision and the perceptual DAME metric. Other methods considerably sacrifice one of these in favour of another.}
{%
\newcolumntype{C}{>{\centering\arraybackslash}X}
\begin{tabularx}{\textwidth}{lCClCC} 
\toprule
\textbf{}     & \multicolumn{2}{c}{\textbf{Training}}       & \textbf{}            & \multicolumn{2}{c}{\textbf{Test}}           \\ \midrule
\textbf{} &
  \begin{tabular}[c]{@{}c@{}}\boldmath{$L_1$} \textbf{Norm}\\ $\mathbf{\times10^{-3} \downarrow}$ \end{tabular} &
  \begin{tabular}[c]{@{}c@{}}\textbf{DAME}\\ $\mathbf{\times10^{-2} \downarrow}$\end{tabular} &
  &
  \begin{tabular}[c]{@{}c@{}}\boldmath{$L_1$} \textbf{Norm}\\ $\mathbf{\times10^{-3} \downarrow}$\end{tabular} &
  \multicolumn{1}{l}{\begin{tabular}[c]{@{}l@{}}\textbf{DAME}\\ $\mathbf{\times10^{-2} \downarrow}$\end{tabular}} \\
  \midrule
\multicolumn{6}{l}{\textbf{Facsimile\texttrademark$ $}}                                                                 \\ \midrule
Ours          & \textbf{1.61}        & \textbf{2.76}        & \multicolumn{1}{c}{} & 6.42                 & 3.17                 \\
DR           & 4.77                 & 3.05                 & \multicolumn{1}{c}{} & 9.29                 & \textbf{3.00}        \\
Euclidean Std. & 2.36                 & 4.90                 & \multicolumn{1}{c}{} & \textbf{5.78}        & 3.89                 \\
Euclidean     & 2.60                 & 6.04                 & \multicolumn{1}{c}{} & 8.32                 & 5.81                 \\ \midrule
              
\multicolumn{6}{l}{\textbf{FaceWarehouse}}                                                             \\ \midrule
Ours          & \hl{\textbf{0.91}}        & \textbf{1.10}       & \textbf{}            & 6.27                 & 1.29                 \\
DR           & 2.20                 & 1.14                 &                      & 7.42                 & \hl{\textbf{1.22}}        \\
Euclidean Std. & 1.11                 & 3.23                 &                      & \hl{\textbf{5.33}}        & 2.53                 \\
Euclidean     & 2.54                 & 5.27                 &                      & 5.34                 & 5.50                 \\ \midrule
             
\multicolumn{6}{l}{\hl{\textbf{FaceScape}}}                                                                   \\ \midrule
Ours          & 1.27                 & 2.25                 & \textbf{}            & 1.65                 & 2.41                 \\
DR           & 6.06                 & \hl{\textbf{1.64}}        &                      & 5.92                 & \hl{\textbf{1.61}}        \\
Euclidean Std. & \hl{\textbf{0.96}}        & 1.81                 &                      & \hl{\textbf{1.30}}        & 1.82                 \\
Euclidean     & 1.32                 & 2.20                 &                      & 1.71                 & 2.26                 \\ \bottomrule
\end{tabularx}%
}
\end{table}

The reconstructions of meshes from the test set reveal a pattern. The standardised Euclidean representation achieves the lowest $L_1$ norm error, and the normalised DR has the lowest DAME error. Nonetheless, in the case of the Facsimile\texttrademark$ $ \cite{HumainLimited2022HumainDevelopment} and \mbox{FaceWarehouse \cite{Cao2014}} datasets, Euclidean and standardised Euclidean coordinates perform the worst on the DAME metric. Analogously, DR has the highest $L_1$ norm. It can be concluded that, with these representations, either point-wise precision or perceptual quality must be sacrificed significantly.

Our method balances these metrics favourably, as demonstrated in Figure \ref{fig:scatter_plots}. The results on the Facsimile\texttrademark$ $ dataset reveal that, compared to normalised DR, our method achieves a 30.9\% lower $L_1$ norm error, with only a 5.7\% increase in DAME error. Contrasted with standardised Euclidean coordinates, our approach yields an 18.5\% lower DAME error at the cost of an 11.5\% higher $L_1$ norm error. Now, turning to the FaceWarehouse \cite{Cao2014} dataset, our method has a 15.5\% lower $L_1$ norm error than normalised DR, with just a 5.7\% higher DAME error. In comparison to standardised Euclidean coordinates, our method results in a 49\% lower DAME error at the cost of just a 17.6\% higher $L_1$ norm error.

However, for the FaceScape \cite{Yang2020} dataset, our method does not improve upon the reconstruction results from other representations on both training and validation sets. Scatter plots in Figure \ref{fig:scatter_plots} reveal the reason. The proposed approach can improve the quality of reconstructed meshes by utilising dedicated mesh representations for each spectral band. Nevertheless, on the FaceScape dataset, the DR representation yields little improvement of the perceptual quality compared to standardised Euclidean representation. Therefore, our proposed method cannot benefit from using different representations in this particular case. Experiments on the FaceScape dataset demonstrate that our method can elevate the overall quality of reconstructed meshes only when there is a substantial difference between the point-wise and the perceptual accuracy yielded by at least two different mesh representations.

Table \ref{table:impact_of_normalisation_and_standardisation} compares the deformation representation (DR) with and without the normalisation preprocessing step described in Section \ref{subsec:high_frequency_band}. Moreover, the table contrasts Euclidean coordinates with and without the standardisation preprocessing step, as described in Section \ref{subsec:low_frequency_band}. Standardisation of low-frequency input features represented in Euclidean coordinates improves training and validation results across all datasets in the comparison. Normalising high-frequency input features represented with DR improves the $L_1$ norm on validation sets and yields a similar or a lower DAME error. The impact of normalising DR is not conclusive on the training dataset, as normalisation of DR improves the $L_1$ norm and DAME metrics on the FaceWarehouse dataset, while DR without normalisation yields better results on the Facsimile\texttrademark$ $ dataset.

\begin{figure}[H]

 \includegraphics[width=0.95\textwidth]{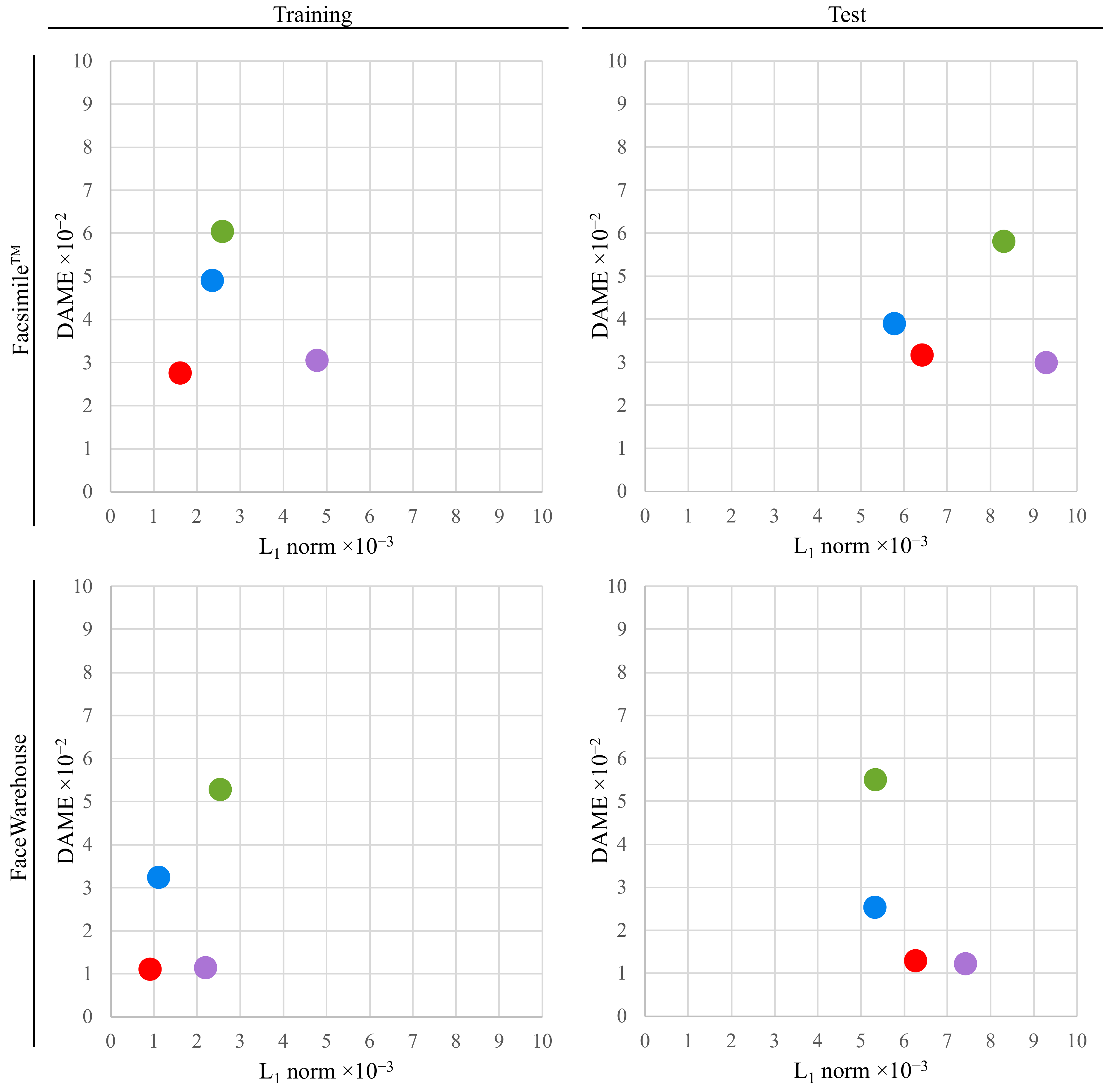}
\caption{\textit{Cont}.}
\end{figure}

\begin{figure}[H]\ContinuedFloat

  \includegraphics[width=0.95\textwidth]{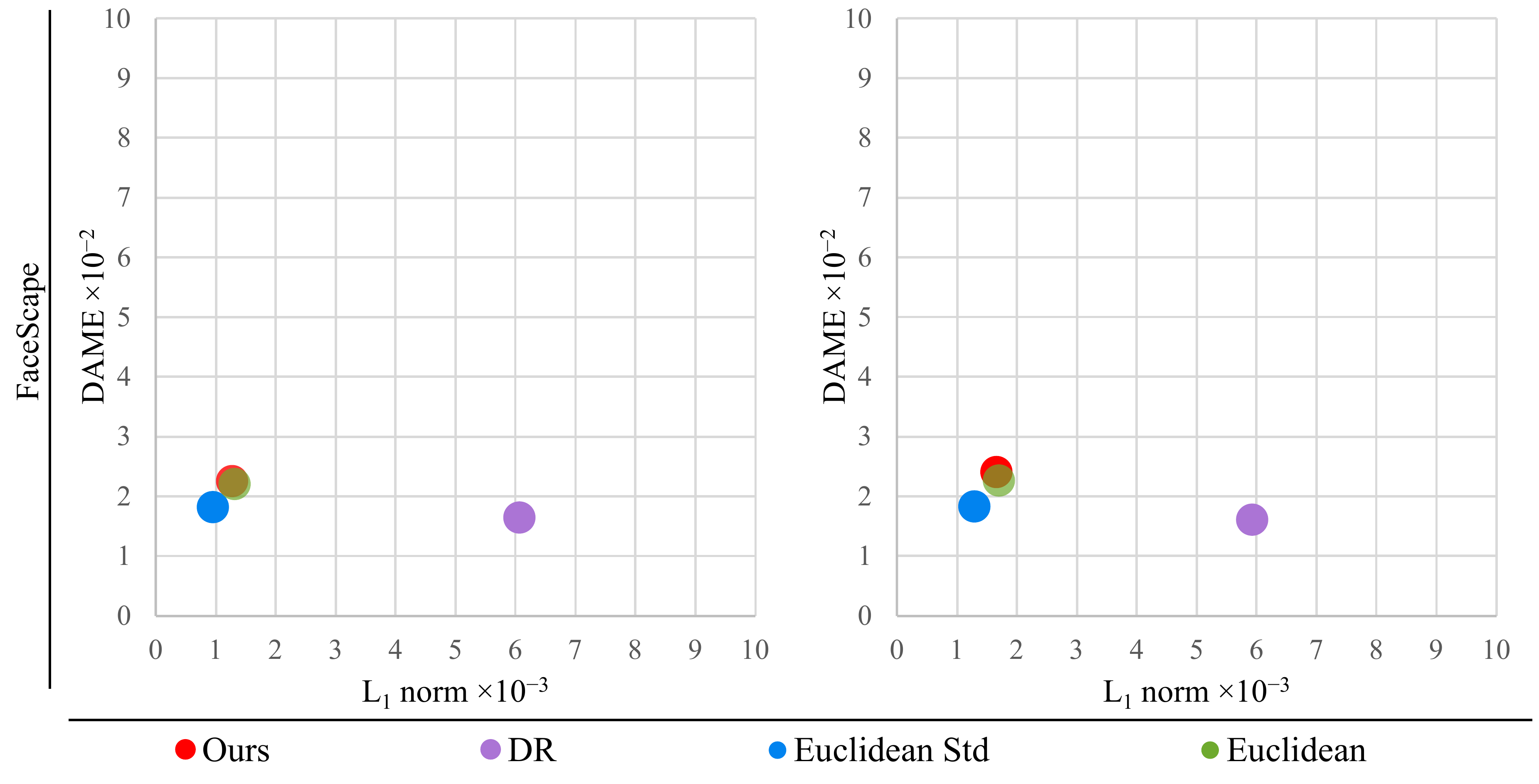}
  \caption{\label{fig:scatter_plots}
          Comparison of the reconstruction results with our method ($k=500$, $\gamma=1$, $\mathbf{Z}=64$) and with common representations used in other methods: Euclidean coordinates \cite{Cheng2019,Hanocka2019MeshCNN:Edge,Zhou2020FullyKernels}, standardised Euclidean coordinates \cite{Bouritsas2019NeuralGeneration,Chen2021LearningModels,Gao2021LearningRepresentation,Gong2019SpiralNet++:Operator,Ranjan2018} and normalised deformation representation (DR) \cite{Jiang2019,Wu2018}. Across Facsimile and FaceWarehouse datasets, our method outperforms in reconstructing examples from the training set and favourably balances perceptual and geometric quality on the Pareto-front of optimal solutions. Our method underperforms on the FaceScape \cite{Yang2020} dataset because the benefit of using normalised DR representation for high-frequency information is minuscule compared to standardised Euclidean representation.}
    
\end{figure}

 \begin{table}[H]

\newcolumntype{C}{>{\centering\arraybackslash}X}
 \caption{\label{table:impact_of_normalisation_and_standardisation} \hl{Ablation study} on the reconstruction task demonstrating the impact of normalisation of the deformation representation (DR) and the standardisation of Euclidean coordinate input features.}
\begin{tabularx}{\textwidth}{lCClCC} 
\toprule
\textbf{}     & \multicolumn{2}{c}{\textbf{Training}}       & \textbf{}            & \multicolumn{2}{c}{\textbf{Validation}}           \\ \midrule
\textbf{} &
  \begin{tabular}[c]{@{}c@{}}\boldmath{$L_1$} \textbf{Norm}\\ $\mathbf{\times10^{-3} \downarrow}$ \end{tabular} &
  \begin{tabular}[c]{@{}c@{}}\textbf{DAME}\\ $\mathbf{\times10^{-2} \downarrow}$\end{tabular} &
  &
  \begin{tabular}[c]{@{}c@{}}\boldmath{$L_1$} \textbf{Norm}\\ $\mathbf{\times10^{-3} \downarrow}$\end{tabular} &
  \multicolumn{1}{l}{\begin{tabular}[c]{@{}l@{}}\textbf{DAME}\\ $\mathbf{\times10^{-2} \downarrow}$\end{tabular}} \\
  \midrule
\multicolumn{6}{l}{\hl{\textbf{Facsimile\texttrademark$ $}}}                                                                  
\\ \midrule
DR without normalisation         & \hl{\textbf{4.01}}        & \hl{\textbf{2.77}}        & \multicolumn{1}{c}{} & 12.05                 & 3.07                 \\
DR with normalisation            & 4.77                 & 3.05                 & \multicolumn{1}{c}{} & \hl{\textbf{9.29}                } & \hl{\textbf{3.00}}        \\
Euclidean without standardisation& 2.60                 & 6.07                 & \multicolumn{1}{c}{} & 8.32        & 5.81                 \\
Euclidean with standardisation & \hl{\textbf{2.36}  }               & \hl{\textbf{4.90}}                 & \multicolumn{1}{c}{} & \hl{\textbf{5.78}}        & \hl{\textbf{3.89} }                \\ 
\midrule
\multicolumn{6}{l}{\hl{\textbf{FaceWarehouse}}}                                                              \\ \midrule
DR without normalisation         & 2.57        & 1.71        &  \multicolumn{1}{c}{}            & 10.03                 & \hl{\textbf{1.19}}                 \\
DR with normalisation            & \hl{\textbf{2.21}}                 & \hl{\textbf{1.14}  }               & \multicolumn{1}{c}{} & \hl{\textbf{7.42}}                 & 1.22        \\
Euclidean without standardisation & 2.54                 & 5.27                 & \multicolumn{1}{c}{} & \hl{\textbf{5.33}}        & 5.50                 \\
Euclidean with standardisation & \hl{\textbf{1.12}  }               & \hl{\textbf{3.23} }                & \multicolumn{1}{c}{} & \textbf{5.33}        & \textbf{2.53}                 \\ \bottomrule

\end{tabularx}%

\end{table}

\subsubsection{Qualitative Evaluation} \label{subsec:qualitative_evaluation}

\begin{figure}[H]

  \includegraphics[width=0.75\textwidth]{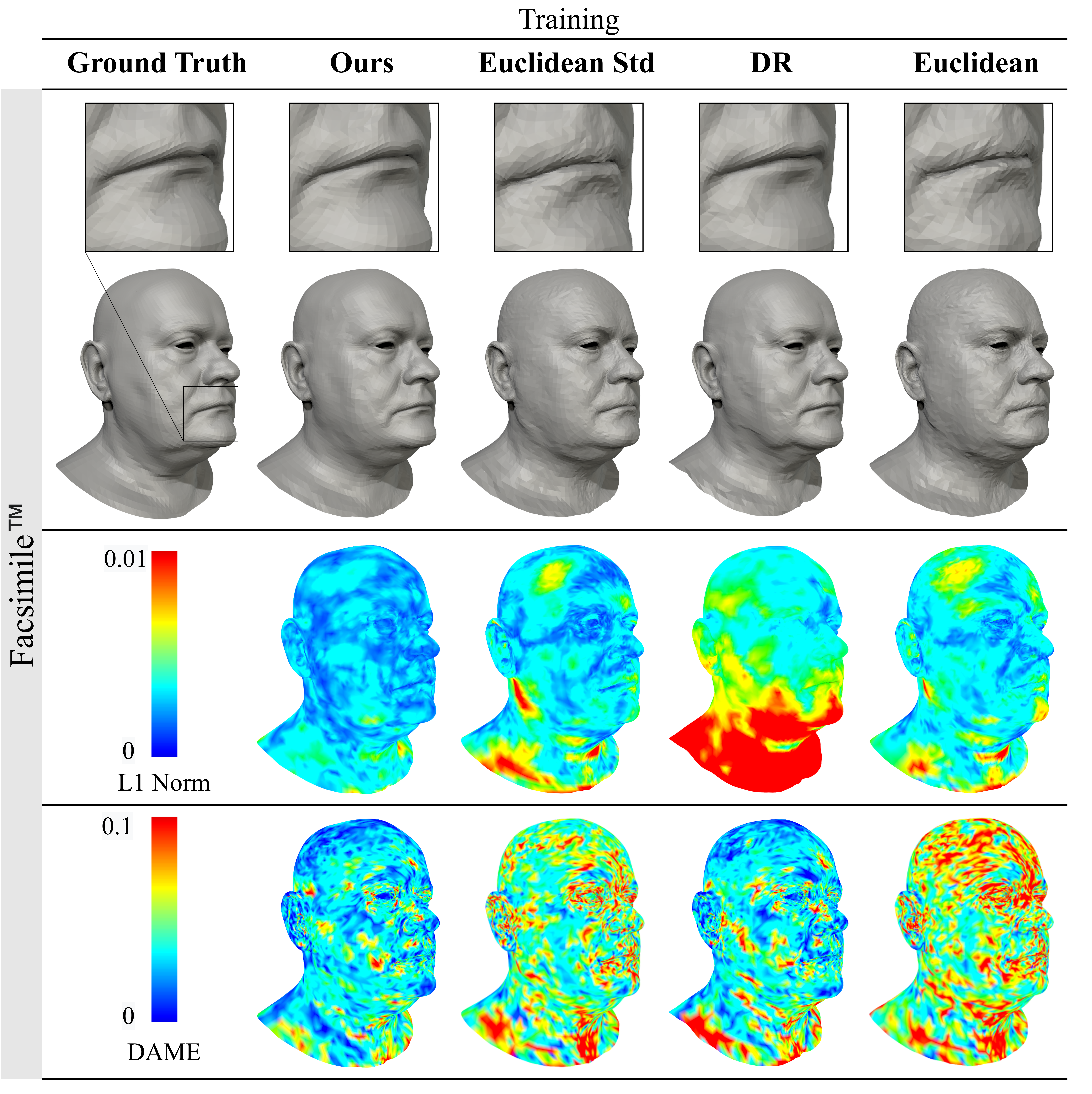}
  \includegraphics[width=0.75\textwidth]{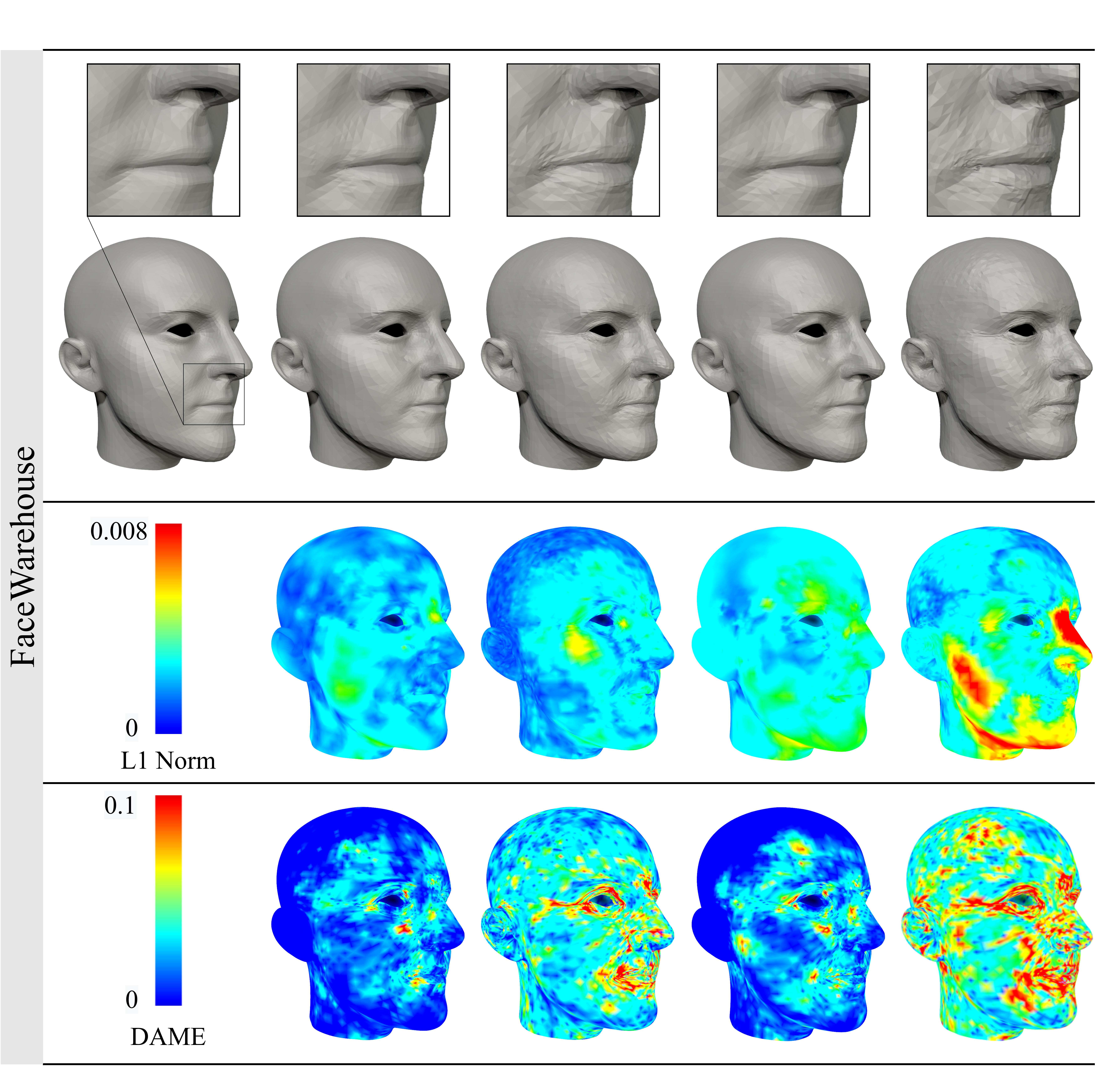}
  \caption{\label{fig:qualitative_reconstruction_train}
          Qualitative comparison of the reconstruction results of training data with our method ($k=500$, $\gamma=1$) and with common representations used in other methods: Euclidean \mbox{coordinates~\cite{Cheng2019,Hanocka2019MeshCNN:Edge,Zhou2020FullyKernels},} standardised Euclidean coordinates \cite{Bouritsas2019NeuralGeneration,Chen2021LearningModels,Gao2021LearningRepresentation,Gong2019SpiralNet++:Operator,Ranjan2018} and the normalised deformation representation (DR) \cite{Jiang2019,Wu2018}. The meshes generated by our method achieve superior results compared to other feature representations. Zooming into the digital version is recommended to see the surface artefacts on the results generated with Euclidean and standardised \mbox{Euclidean representations.}}
    
\end{figure}

\begin{figure}[H]

  \includegraphics[width=0.75\textwidth]{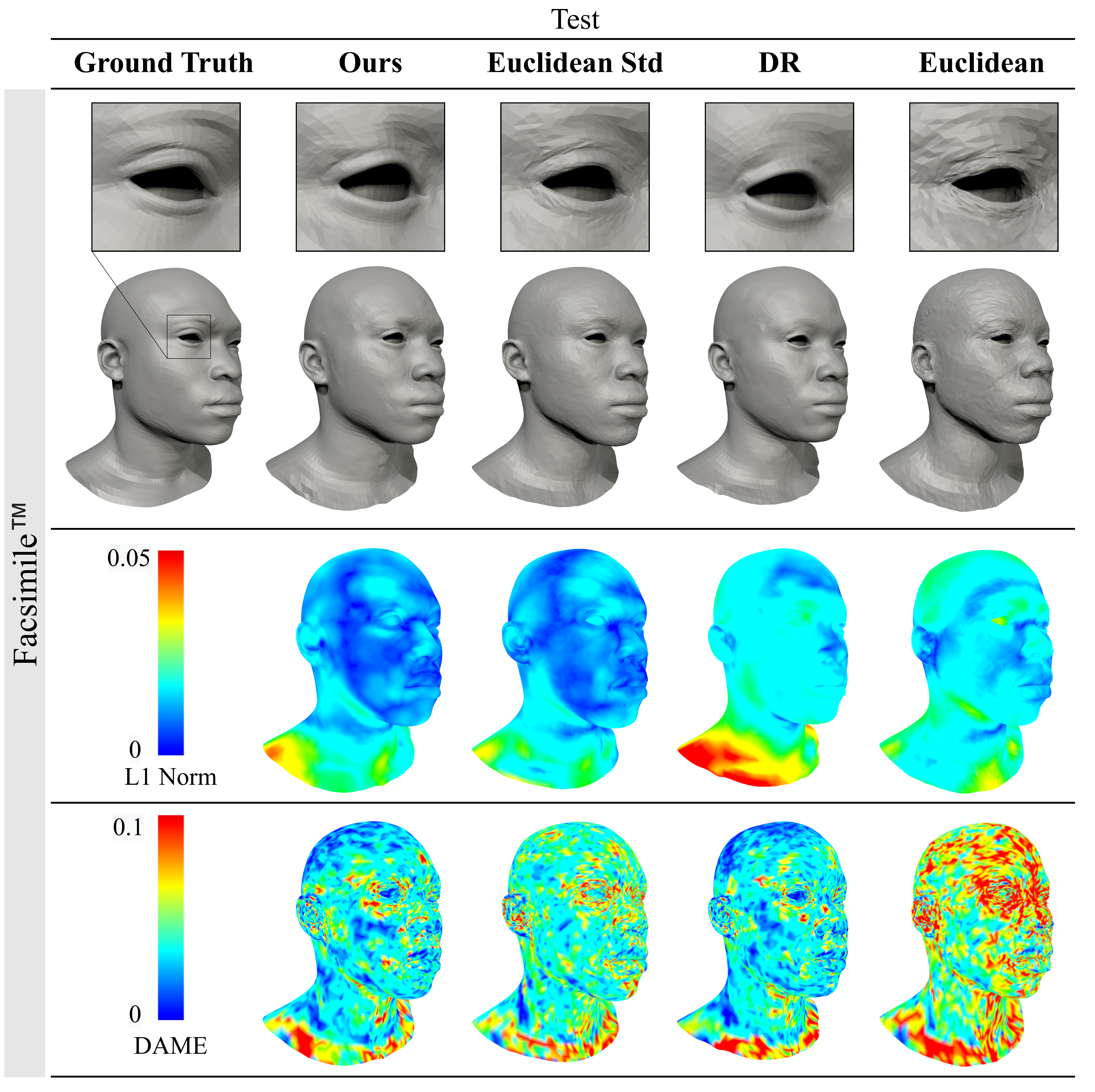}
  \includegraphics[width=0.75\textwidth]{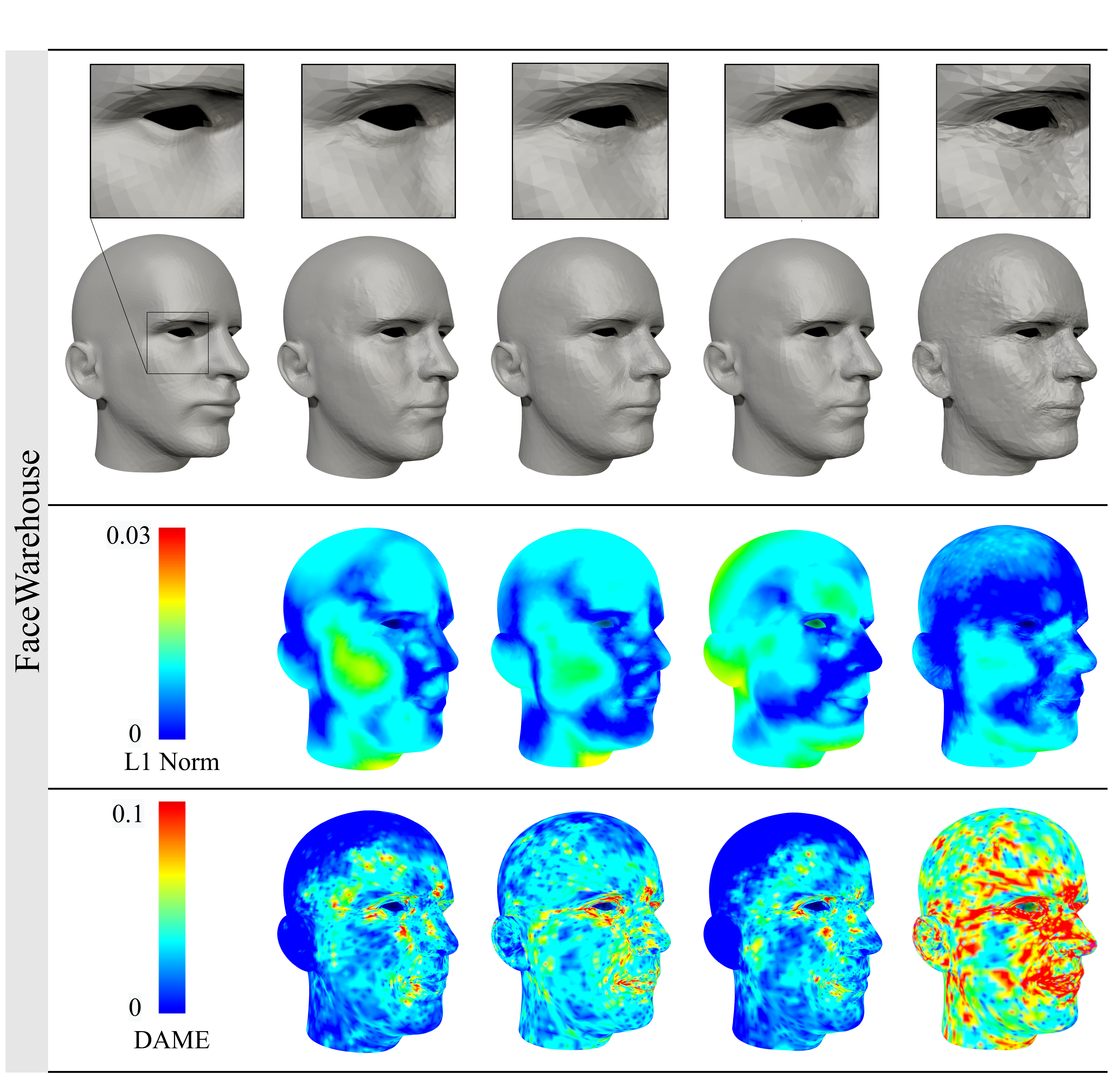}
  \caption{\label{fig:qualitative_reconstruction_test}
          Qualitative comparison of the reconstruction results of test data with our method ($k=500$, $\gamma=1$) and with common representations used in other methods: Euclidean coordinates \cite{Cheng2019,Hanocka2019MeshCNN:Edge,Zhou2020FullyKernels}, standardised Euclidean coordinates \cite{Bouritsas2019NeuralGeneration,Chen2021LearningModels,Gao2021LearningRepresentation,Gong2019SpiralNet++:Operator,Ranjan2018} and the normalised deformation representation (DR) \cite{Jiang2019,Wu2018}. The meshes generated by our method have similar surface quality to the outputs with DR while achieving much lower volume loss in the neck, chin, and cheek areas. \newline}
    
\end{figure}

Figures~\ref{fig:qualitative_reconstruction_train} and \ref{fig:qualitative_reconstruction_test} provide a visual assessment of the mesh quality in the reconstruction experiments, comparing our method with other commonly used representations. The meshes generated using Euclidean and standardised Euclidean representations display visible surface artefacts and struggle to capture high-frequency details. The severity of surface discrepancies corresponds with the colour visualisation of the DAME error. Results obtained with the normalised deformation representation (DR) are perceptually more similar to the ground truth meshes. Nonetheless, there is noticeable volume loss in the neck, chin and cheeks. In contrast, our method produces results that are perceptually similar to ground truth meshes without experiencing noticeable volume changes.

\unskip
\unskip
 
\begin{figure}[H]

  \includegraphics[width=0.9\textwidth]{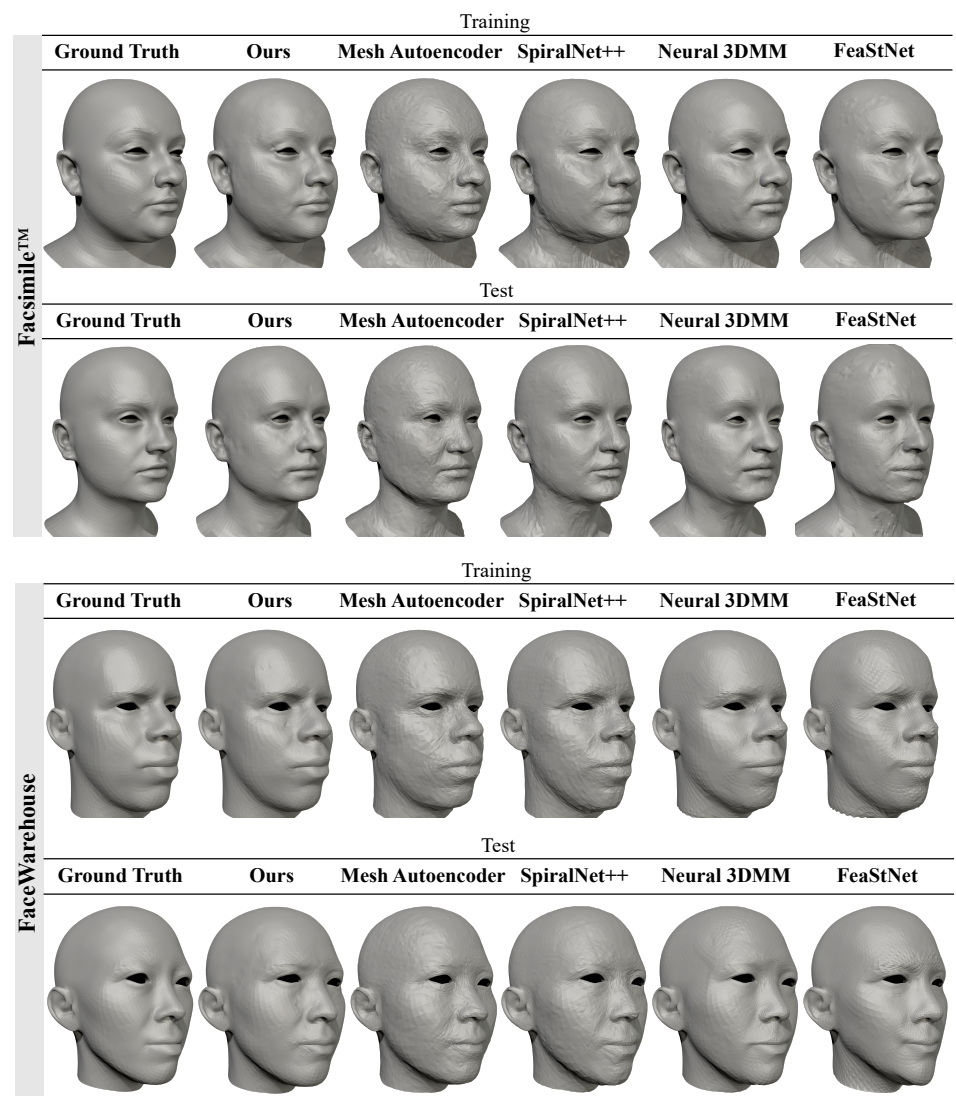}
  \caption{\label{fig:visual_comparison_to_other_methods}
          Visual comparison of the reconstruction results of the Facsimile\texttrademark$ $ and FaceWarehouse datasets using our method ($k=500$, $\gamma=1$, $\mathbf{Z}=64$) and four other methods: Mesh Autoencoder \cite{Zhou2020FullyKernels}, SpiralNet++ \cite{Gong2019SpiralNet++:Operator}, Neural 3DMM \cite{Bouritsas2019NeuralGeneration} and FeaStNet \cite{Verma2018}. It is recommended to zoom into the digital version to compare the reconstructed meshes.}
\end{figure}

Figure~\ref{fig:visual_comparison_to_other_methods} compares the rendered meshes from the Facsimile\texttrademark$~$ and FaceWarehouse datasets reconstructed with our proposed method and other popular methods: Mesh Autoencoder \cite{Zhou2020FullyKernels}, SpiralNet++ \cite{Gong2019SpiralNet++:Operator}, Neural 3DMM \cite{Bouritsas2019NeuralGeneration} and FeaStNet \cite{Verma2018}. A user study was conducted to qualitatively assess the perceptual quality of meshes generated by our method ($k=500$, $\gamma=1$, $\mathbf{Z}=64$) compared with meshes synthesised using alternative methods and common feature representations. The evaluation focused on reconstructions of training and test subsets of the Facsimile\texttrademark$~$ and FaceWarehouse datasets. The study was conducted online on a representative sample of 94 participants (42 female, 51 male, 1 non-binary), with the following age distribution: 25 participants aged 18--25, 52 aged 26--35, 11 aged 36--45 and 6 aged 46--55. Participants viewed three images of rendered 3D models and were instructed to compare the ground truth reference model (always in the middle) against models A (left) and B (right). Subsequently, participants selected the model they perceived as more similar to the reference model (either option ``A'' or ``B''). In cases where a clear distinction was challenging, participants had the option to choose ``Difficult to say''. Each participant completed a total of 48 comparisons. In each comparison, one 3D model was generated by our method, while the other was produced by an alternative method or using a different commonly used representation. Participants were instructed to use a desktop monitor or a tablet and complete the study in a full-screen view.

\begin{figure}[H]
  \centering

  \includegraphics[width=\textwidth]{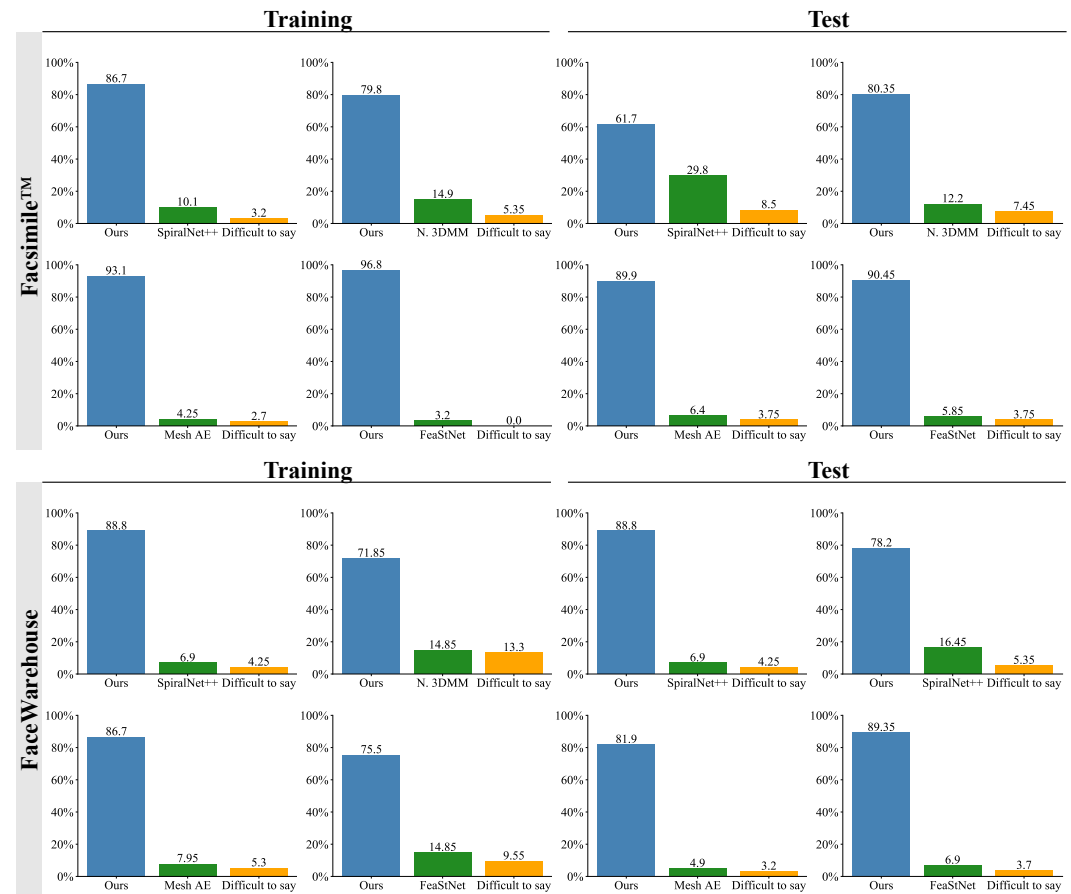}
  \caption{\label{fig:user_study_methods} \hl{The outcomes}
  of the user study, which compared the visual similarity to the ground truth of the meshes generated by our method and other methods: Mesh Autoencoder \cite{Zhou2020FullyKernels}, SpiralNet++ \cite{Gong2019SpiralNet++:Operator}, Neural 3DMM \cite{Bouritsas2019NeuralGeneration} and FeaStNet \cite{Verma2018}. The bars show the percentage of participants who selected the mesh generated by the given method as more similar to the ground truth mesh. The participants were asked to select "Difficult to say" only when they had to guess between the generated models.}
    
\end{figure}

Figure~\ref{fig:user_study_methods} displays the aggregated responses from the user study, which compared the visual similarity of the meshes generated by our proposed method with those produced by other methods \cite{Zhou2020FullyKernels, Gong2019SpiralNet++:Operator,Bouritsas2019NeuralGeneration,Verma2018}. In all cases, a strong majority of participants perceived the meshes generated by our method as more similar to the reference than those generated by other compared methods. Participants expressed a high certainty of their responses, with only a median of 4.3\% choosing the ``Difficult to say'' option.
    
\begin{figure}[H]

  \includegraphics[width=\textwidth]{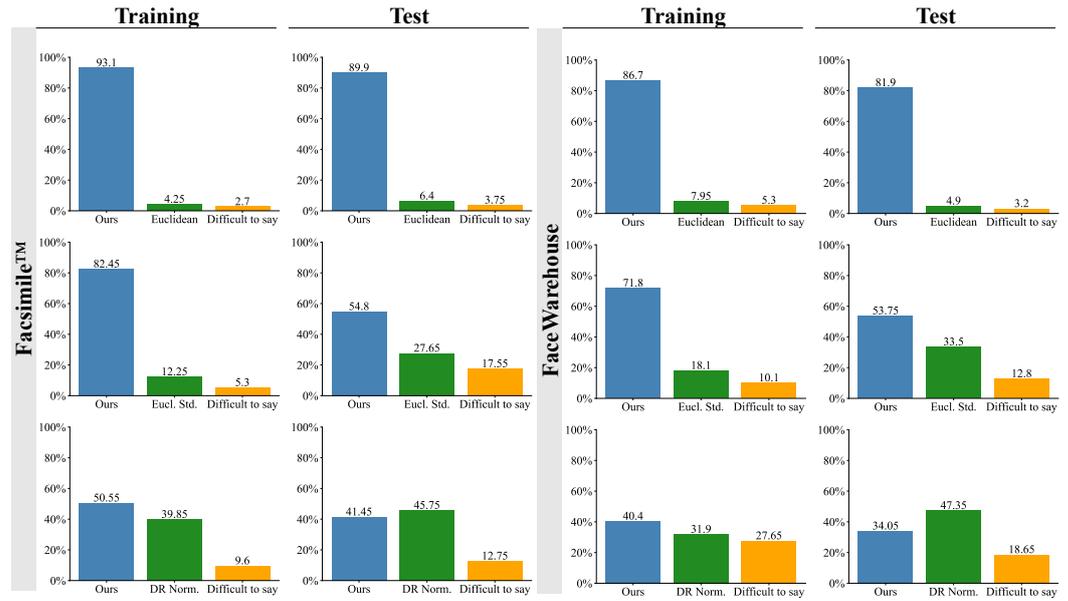}
  \caption{\label{fig:user_study_representations} \hl{The results}
  from the user study comparing our method with common representations used in other methods: Euclidean coordinates \cite{Cheng2019,Hanocka2019MeshCNN:Edge,Zhou2020FullyKernels}, standardised Euclidean coordinates \mbox{(Eucl. Std.)}~\cite{Bouritsas2019NeuralGeneration,Chen2021LearningModels,Gao2021LearningRepresentation,Gong2019SpiralNet++:Operator,Ranjan2018} and the normalised deformation representation (DR Norm.) \cite{Jiang2019,Wu2018}. The bars show the percentage of participants who selected the mesh generated by the given method as more similar to the ground truth mesh. The participants were asked to select ``Difficult to say'' only when they had to guess between the generated models.}
    
\end{figure}

Figure~\ref{fig:user_study_representations} presents the results from the user study comparing our method with common representations used in other methods: Euclidean coordinates \cite{Cheng2019,Hanocka2019MeshCNN:Edge,Zhou2020FullyKernels}, standardised Euclidean coordinates \cite{Bouritsas2019NeuralGeneration,Chen2021LearningModels,Gao2021LearningRepresentation,Gong2019SpiralNet++:Operator,Ranjan2018} and the normalised deformation representation \mbox{(DR) \cite{Jiang2019,Wu2018}.} Participants' responses align with our quantitative analysis of perceptual DAME error from Table~\ref{table:quantitative_results}. On average, 87.9\% and 65.7\% of participants perceived our methods' results as more similar than those from methods using Euclidean coordinates and standardised Euclidean coordinates, respectively. Regarding the comparison with normalised DR representation, the participants were divided. The training sets reconstructed with our method received 10.7 and 8.5 more percentage points of participants' preference. In comparison, the meshes produced with the normalised DR representation were chosen as more similar on the test sets by 4.3 and 13.3 more percentage points of participants. Consequently, the perceptual similarity of meshes generated by our method and the normalised DR representation is comparable, while our method yields significantly lower point-wise accuracy error, as demonstrated in quantitative analysis.

\subsection{Mesh Interpolation}

Mesh interpolation is widely applied in facial animation to produce new facial meshes from two known facial meshes. In contrast to existing interpolation methods that interpolate two facial meshes, our proposed approach interpolates low- and high-frequency parts of two facial meshes.

\begin{figure}[H]
  \centering

  \includegraphics[width=\textwidth]{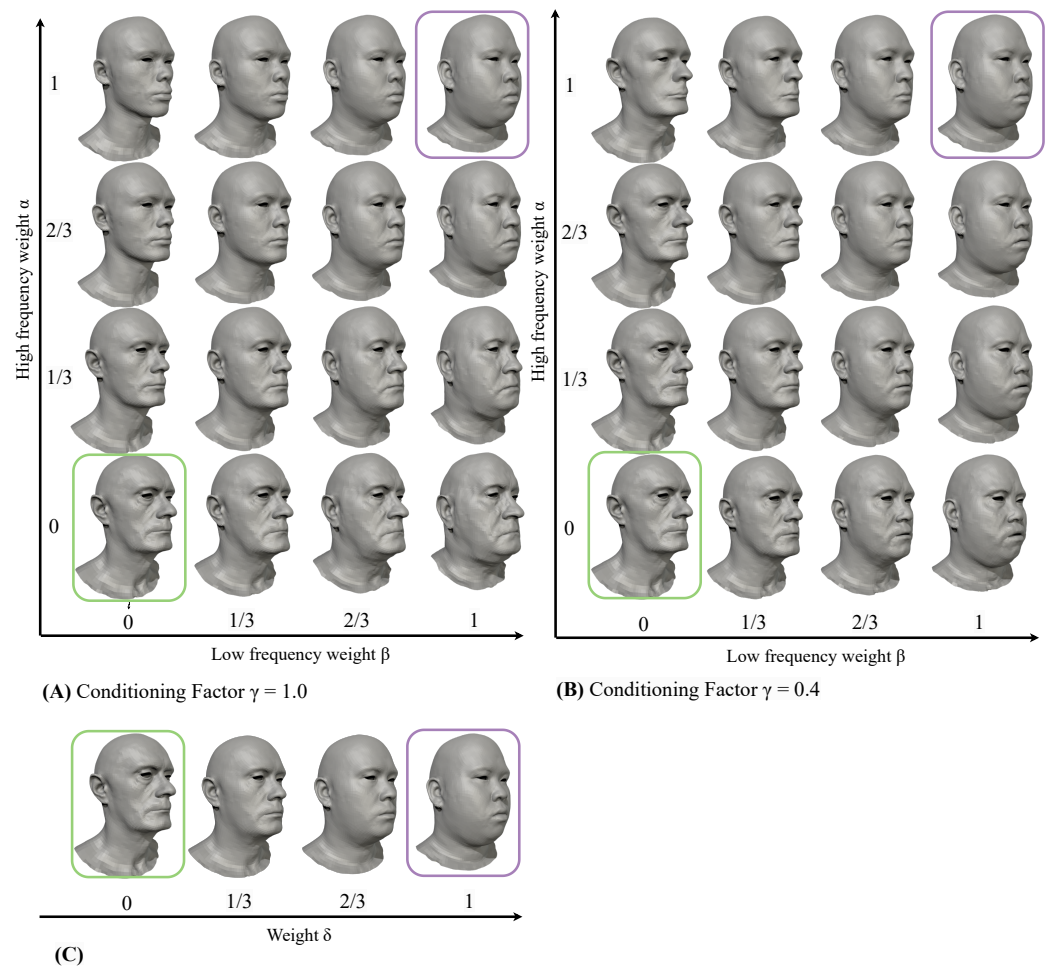}
  \caption{\label{fig:interpolation}
          Interpolation of low-frequency and high-frequency latent parameters, $k=500$. Two facial meshes (in green and purple outlines) from the Facsimile\texttrademark$ $ \cite{HumainLimited2022HumainDevelopment} dataset are encoded. In (\textbf{A}), the model is trained with the Conditioning Factor $\gamma=1.0$. In (\textbf{B}), the Conditioning Factor $\gamma=0.4$. The meshes arranged in a grid are decoded from interpolated latent parameters. In (\textbf{C}), the meshes in green and purple outlines are interpolated in the vertex space.}
    
\end{figure}

In Figure \ref{fig:interpolation}, the subject in the green outline is encoded into parameters $[\mathbf{Z}_{1l} | \mathbf{Z}_{1h}]$ and the subject in the purple outline is encoded into parameters $[\mathbf{Z}_{2l} | \mathbf{Z}_{2h}]$. High-frequency weight $\alpha$ and low-frequency weight $\beta$ are used to interpolate between these latent parameters so that the interpolated latent parameters are
\begin{equation} \label{eqn:interpolation}
\begin{aligned}
    \mathbf{Z}_{\alpha, \beta} = [(1-\beta)\mathbf{Z}_{1l} + \beta\mathbf{Z}_{2l} | (1-\alpha)\mathbf{Z}_{1h} + \alpha \mathbf{Z}_{2h}].
\end{aligned}
\end{equation}

The meshes resulting \hl{from}
Equation (\ref{eqn:interpolation}) are shown in Figure~\ref{fig:interpolation}A,B. In the figure, the meshes arranged in a grid are decoded from latent parameters $\mathbf{Z}_{\alpha, \beta}$. When using the linear interpolation in the vertex space between the mesh in the green outline and the mesh in the purple outline, the interpolation equation is
\begin{equation} \label{eqn:linear_interpolation}
\begin{aligned}
    \mathbf{P}_{\delta} = \mathbf{P}_{1} + \delta(\mathbf{P}_{2}-\mathbf{P}_{1})   ,
\end{aligned}
\end{equation}
where $\mathbf{P}_{1}$ and $\mathbf{P}_{2}$ are the vertex coordinates of the meshes in the green outline and the purple outline, respectively, and $0\le\delta\le1$.

The meshes obtained from Equation (\ref{eqn:linear_interpolation}) are shown in Figure~\ref{fig:interpolation}C. Interpolating low- and high-frequency latent parameters with two different conditioning values, 0.4 and 1.0, generates 28 new meshes. In contrast, interpolating the meshes in the green and purple outlines creates only two new meshes.

Our discussion indicates that multi-frequency interpolation noticeably raises the capacity to create novel facial meshes from two known facial meshes. In addition, Figure \ref{fig:interpolation} demonstrates the disentanglement of low and high frequencies in the parametric space and illustrates the impact of the Conditioning Factor $\gamma$.

When $\gamma=1.0$, the high-frequency parameters of an elderly subject influence mid-frequencies to impose the generation of plausible faces. However, due to bias towards younger subjects in the dataset, this conditioning significantly restricts the domain of generated faces. For an artist, such editing behaviour may be undesirable. In contrast, interpolation of low- and high-frequency parameters with $\gamma=0.4$ is more predictable, as high- and low-frequency deformations are almost fully disentangled. Nevertheless, plausible faces exist within a joint distribution of low- and high-frequency deformations. Consequently, the network may generate implausible examples, like the older man with young, smooth skin depicted in Figure \ref{fig:interpolation}B at ($\alpha=1, \beta=0$). Therefore, the choice of $\gamma$ depends on application-specific requirements, whether prioritising more precise and flexible artistic control or the ability to generate only plausible faces.

\subsection{Multi-Frequency Editing}

\begin{figure}[H]

  \includegraphics[width=\textwidth]{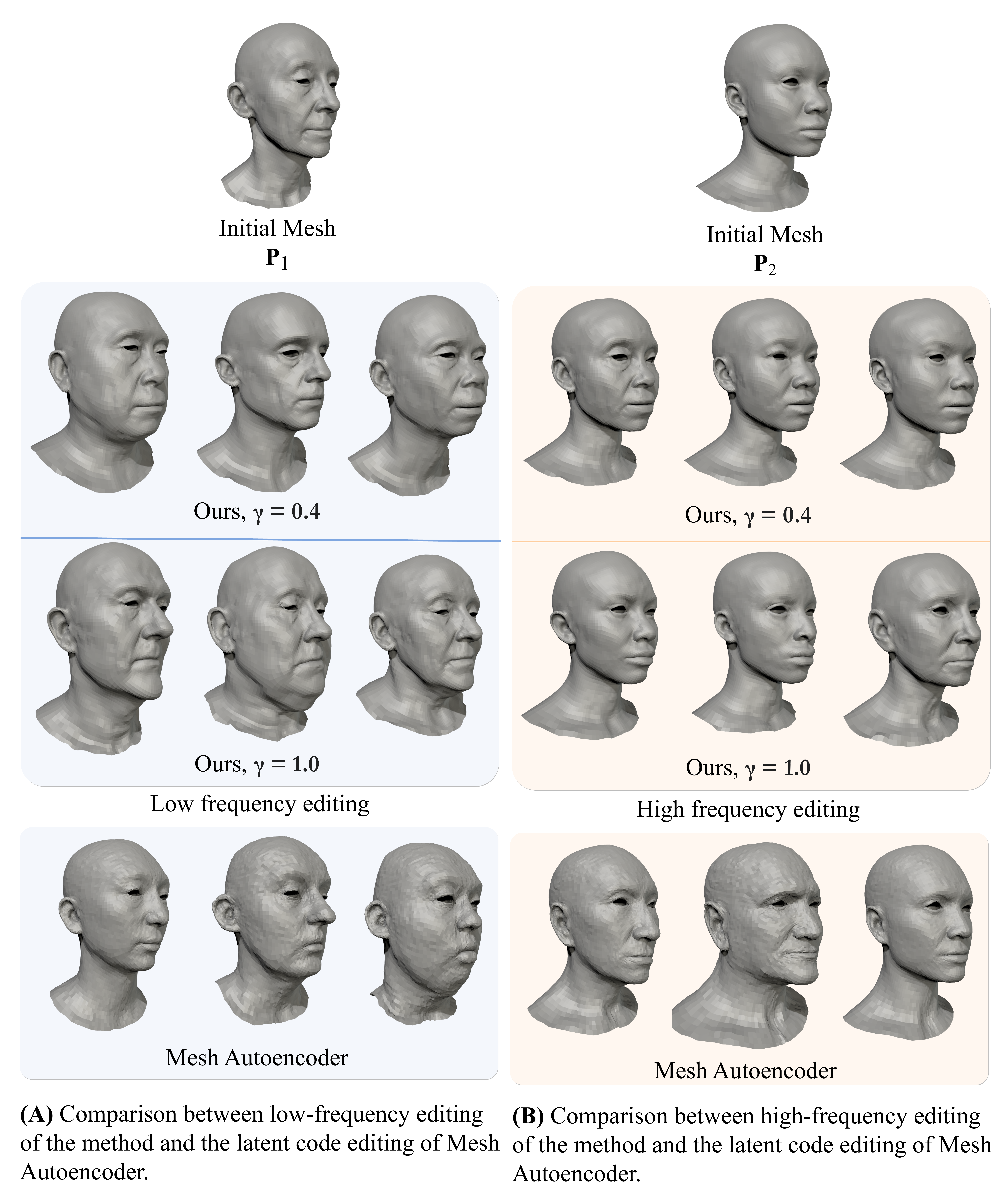}
  \caption{\label{fig:editing}
          Comparison of latent code editing between the proposed method and Mesh \mbox{Autoencoder \cite{Zhou2020FullyKernels}.} In (\textbf{A}), the editing of low-frequency latent codes of encoded mesh $\mathbf{P}_1$. In (\textbf{B}), the editing of high-frequency latent codes of encoded mesh $\mathbf{P}_2$. Top and middle row: the examples decoded using our model with $k=500$ and Conditioning Factor $\gamma=0.4$ and $\gamma=1.0$. Bottom row: the results of editing a subset of latent parameters using the method in \cite{Zhou2020FullyKernels}. The parameters of our method successfully disentangle high and low frequencies. While subjective, it can be observed that lower $\gamma$ provides more control and produces more diverse results. Meanwhile, altering the parameters of Mesh Autoencoder \cite{Zhou2020FullyKernels} affects the entire frequency spectrum.}
    
\end{figure}

Figure \ref{fig:editing} presents the application of our method in editing low- and high-frequency deformations independently. Additionally, the editing capabilities of our model are compared with the method in \cite{Zhou2020FullyKernels}, which does not disentangle low- and high-frequency parameters. In our approach, low-frequency parameters affect the overall head shape while preserving high-frequency details. The Conditioning Factor $\gamma$ impacts the nature of editing. With $\gamma = 1.0$ (full conditioning), high frequencies condition low-frequency deformations to a more narrow domain of only plausible faces. In contrast, when $\gamma = 0.4$ (partial conditioning), edited heads are more diverse, despite some falling outside of the domain of real faces. When editing high-frequency parameters, the overall head shape remains unchanged, with only fine details being affected. Notably, with $\gamma=0.4$, the editing of high frequencies becomes more precise and predictable since high- and low-frequency parameters barely influence each other.

\section{Summary and Future Work} \label{sec:conclusions}
In this paper, a new approach called Deep Spectral Meshes was proposed to address two crucial challenges: the absence of dedicated representations for describing deformations at different frequencies and the inability to independently edit deformations at different frequency levels. These problems may lead to semantically meaningless parameters used in deep learning and result in the unsatisfactory geometric and perceptual quality of assets failing to meet standards in industrial applications.

To develop our new approach, spectral meshes were introduced, overcoming these challenges. They were employed to decompose mesh deformations into low- and high-frequency parts. Subsequently, these parts were converted into features represented with standardised Euclidean coordinates and the normalised deformation representation, respectively. Graph neural networks were proposed to reconstruct features, which were later converted back to Euclidean coordinates to obtain the reconstructed 3D models. A Conditioning Factor was introduced to control the level of mutual conditioning of deformations at different frequencies.

Extensive experiments were conducted to validate and compare our proposed approach with previously published methods, both quantitatively and qualitatively. Results demonstrate the capability of our approach to independently edit low- and high-frequency deformations of facial meshes. Moreover, the experiments illustrate the impact of the Conditioning Factor on balancing mutually exclusive objectives of independent control of deformations at different frequencies and generating plausible synthetic examples. Comparisons with existing methods demonstrate the superiority of our approach over Euclidean coordinates, standardised Euclidean coordinates, the normalised deformation representation (DR) and other methods, as assessed by both $L_1$ and perceptual metric evaluations.

The significance of our proposed approach is underscored through its applications presented in this paper. As described in Section \ref{sec:results}, our method has been applied in mesh compression, enhancing the quality of the reconstructed meshes. Additionally, it has been employed in mesh interpolation, expanding the capability to generate a larger variety of new shapes compared to direct interpolation between two facial meshes. This was achieved by independently interpolating both low-frequency and high-frequency parameters. Moreover, our proposed method has found application in multi-frequency editing, satisfying different editing requirements. It allows the alteration of the overall shape while preserving details by adjusting only high-frequency parameters. It also accommodates modifying fine details while maintaining the overall shape by editing only low-frequency parameters.

\textls[-8]{Applications of Deep Spectral Meshes are potentially far-reaching, but we identified several promising applications and areas of research which warrant \mbox{further investigation:}}
\begin{itemize}[leftmargin=*, labelsep=4.9mm] 
    \item This work restricts its application to low- and high-frequency bands. The partition of mesh data into more than two frequency bands and an investigation into the relationship between the learning model and the frequency bands remain unexplored. The limitation to two frequency bands in our approach is linked to the properties of the chosen mesh representations. Standardised Euclidean coordinates represent low-frequency information to ensure high point-wise accuracy of the generated meshes. The normalised deformation representation (DR) encodes high-frequency information for superior perceptual quality of the results. Future work on partitioning mesh data into more than two frequency bands could further explore the benefits of alternative feature representations at different frequency levels.
    
    \item Further research could describe the relationship between the choice of parameter $k$ and the quality of generated meshes. Designing an algorithm to determine the optimal split between frequency bands is a potential avenue for future research. To achieve this, an objective function relating the quality of generated meshes to parameter $k$ could be formulated. The selection of a suitable optimisation method would be crucial to efficiently obtain the optimal split.
    
    \item Our proposed method can be extended to address the problem of multi-frequency-based deformation transfer, which has not been investigated in existing research studies. The basic idea of multi-frequency-based deformation transfer involves decomposing source and target meshes into mean, low- and high-frequency parts. The differences between the source model and these parts at two different poses are determined and transferred to the corresponding bands of the target mesh. Subsequently, the graph neural network proposed in this paper can be employed to reconstruct a new shape for the target mesh with the pose of the source mesh.
    
    \item While our approach has been applied to deformable facial meshes, future work could extend the proposed method to articulated shapes like hands and bodies. Existing research has proposed various methods to relate articulated shapes to their underlying skeleton. With this extension, articulated shapes can be decomposed into mean, low- and high-frequency parts. Then, the relationships between these parts and the movements of the skeleton of the articulated shapes can be investigated. These relationships can be used to synthesise mean, low- and high-frequency parts of new poses. The graph neural network proposed in this paper can then extract features, reconstruct them, and synthesise new shapes from the reconstructed features.

    \item It may be possible to apply the approach proposed in this paper to 3D tumour image analysis. First, 3D shapes containing tumours can be reconstructed from medical images. Then, the approach could be employed to disentangle normal deformations and abnormal deformations caused by tumours and extract tumour features.
\end{itemize}

\vspace{6pt} 

\authorcontributions{Conceptualisation, methodology, software, validation, visualisation, writing---original draft preparation, R.K.; writing---review and editing, R.K., R.S., L.Y., S.B., W.K. and G.M.; supervision, R.S., L.Y., W.K. and G.M.; resources and data curation, W.K. All authors have read and agreed to the published version of the manuscript.}

\funding{This work was supported by grant funding of the Centre for Digital Entertainment at Bournemouth University by the UK Engineering and Physical Sciences Research Council (EPSRC) EP/L016540/1 and Humain Ltd.}

\dataavailability{The FaceWarehouse \cite{Cao2014} dataset is available from Kun Zhou, Zhejiang University. Data are available at  \url{http://kunzhou.net/zjugaps/facewarehouse/} \hl{accessed on 2 October 2020}
 with the permission of Kun Zhou. The FaceScape \cite{Yang2020} dataset is available from the CITE Lab, Nanjing University. Data are available at  \url{https://facescape.nju.edu.cn/} \hl{accessed on 3 July 2020} with the permission of the CITE Lab. Both of the datasets are available to approved researchers due to the sensitive nature of human face data. The datasets were used under licenses for this study. For commercial reasons, the Facsimile\texttrademark$ $ \cite{HumainLimited2022HumainDevelopment} dataset is not publicly available.}

\conflictsofinterest{R.K., S.B., W.K. and G.M. are employees of Humain Ltd. R.S. is an employee of The Foundry Visionmongers Ltd. The authors declare no conflicts of interest.}

\begin{adjustwidth}{-\extralength}{0cm}

\reftitle{References}

\PublishersNote{}
\end{adjustwidth}
\end{document}